\pdfoutput=1

\documentclass[letterpaper,twocolumn,10pt]{article}
\usepackage{usenix2019_v3}

\usepackage{cite}
\usepackage{amsmath,amssymb,amsfonts}
\usepackage{algorithmic}
\usepackage{graphicx}
\usepackage{epstopdf}
\usepackage{textcomp}
\usepackage{xcolor}
\usepackage{booktabs}
\usepackage{threeparttable}
\usepackage{multirow}
\usepackage{bigstrut}
\usepackage{subfigure}
\usepackage{array}
\usepackage{algorithm}
\usepackage{enumerate}

\usepackage{float}
\usepackage{threeparttable}
\usepackage{url}
\usepackage{booktabs}
\usepackage{tikz}
\usepackage{amsmath}
\usepackage{verbatimbox}
\usepackage[numbers,sort&compress]{natbib}
\usepackage{caption}
\usepackage{setspace}
\usepackage{filecontents}

\newcommand{\tabincell}[2]{\begin{tabular}{@{}#1@{}}#2\end{tabular}}

\begin{document}

\title{3D Invisible Cloak}

\author{
{\rm Mingfu~Xue, Can He, Zhiyu Wu, Jian Wang, Zhe Liu, and Weiqiang~Liu }\\
  Nanjing University of Aeronautics and Astronautics\\
  \{mingfu.xue, hecan, wuzhiyu, wangjian, zhe.liu, liuweiqiang\}@nuaa.edu.cn
}%

\maketitle
\begin{abstract}
Deep neural network (DNN) based object detectors have been widely deployed in many areas. However, recent studies show that DNN model is vulnerable to various attacks, \textit{e.g.}, adversarial example attacks. The perturbations in the input instances can cause the DNN models to output incorrect predictions, which will pose serious threats to security or safety critical applications.
In this paper, we propose a novel physical stealth attack against the person detectors in real world.
The proposed method generates an adversarial patch, and prints it on real clothes to make a three dimensional (3D) invisible cloak.
Anyone wearing the cloak can evade the detection of person detectors and achieve stealth.
For the first time, we consider the impacts of those 3D physical constraints (\textit{i.e.}, radian, wrinkle, occlusion, angle, \textit{etc}.) on person stealth attacks, and propose 3D transformations to generate 3D invisible cloak. We launch the person stealth attacks in 3D physical space instead of 2D plane by printing the adversarial patches on real clothes under challenging and complex 3D physical scenarios.
The conventional and 3D transformations are performed on the patch during its optimization process, to make the generated 3D invisible cloak robust in realistic scenes.
Further, we study how to generate the optimal 3D invisible cloak. Specifically, we explore how to choose input images with specific shapes and colors to generate the optimal 3D invisible cloak, and draw some important conclusions.
Besides, after successfully making the object detector misjudge the person as other objects, we explore how to make a person completely disappeared, \textit{i.e.}, the person will not be detected as any objects.
Finally, we present a systematic evaluation framework to methodically evaluate the performance of the proposed attack in digital domain and physical world.
Experimental results in various indoor and outdoor physical scenarios show that, the proposed person stealth attack method is robust and effective even under those complex and challenging physical conditions, such as the cloak is wrinkled, obscured, curved, and from different angles.
The attack success rate of the generated adversarial patch in digital domain (Inria data set) is 86.56\%, while the static and dynamic stealth attack performance of the generated 3D invisible cloak in physical world is 100\% and 77\%, respectively, which are significantly better than existing works.
\end{abstract}

\vspace{-0.25cm}
\section{Introduction}
\vspace{-0.25cm}
In recent years, deep learning has achieved significant technical breakthroughs and been widely applied in many areas.
Deep neural network (DNN) models have applied in various safety or security related applications, such as self-driving cars \cite{geiger2012we}, face detection \cite{ouyang2015deepid}, and intelligent video surveillance \cite{pan2018deep}, \textit{etc}.
However, recent studies show that DNN models are vulnerable to adversarial example attacks in digital domain \cite{carlini2017towards,szegedy2013intriguing,goodfellow2014explaining,papernot2016limitations,su2019one,moosavi2016deepfool} and even in real physical world \cite{kurakin2016adversarial,athalye2018synthesizing,eykholt2018robust,jan2019connecting, yang2018building, thys2019fooling}.

In real world, the attackers may attempt to evade the detection of a person detection system or an intelligent monitoring system.
Compared to these digital adversarial example attacks \cite{carlini2017towards,szegedy2013intriguing,goodfellow2014explaining,papernot2016limitations,su2019one,moosavi2016deepfool}, attacking the detectors in real world are more challenging due to the variabilities of physical conditions \cite{eykholt2018robust}, such as the changes of angles and distances.
These physical constraints will cause the perturbations in an adversarial example be ignored by the detectors after undergoing the physical transformations.
Moreover, the stealth attack against the person detectors has some additional challenges: 1) the individual differences among humans are more obvious \cite{thys2019fooling}, which require the generated adversarial examples to adapt to these intra-class changes; 2) humans are moving constantly, which makes the background interferences more randomized and complicated; 3) humans' postures are changing dynamically, thus the shape of cloak will change from time to time.

So far, there are only two exploratory works on person stealth attacks against the object detectors that deployed in real world \cite{yang2018building, thys2019fooling}.
However, these two works did not take into account the real complex 3D factors during the adversarial examples generation.
Besides, the two works just displayed the generated adversarial stickers on the tablet computer's screen \cite{yang2018building}, or printed the adversarial patch on a rigid cardboard \cite{thys2019fooling}.
This makes the two stealth attacks \cite{yang2018building, thys2019fooling} essentially belong to the two-dimensional (2D) plane attacks.
In addition, the two studies \cite{yang2018building, thys2019fooling} lacked methodical and statistical evaluations for their attacks.
Most experiments in \cite{yang2018building} were simulations in digital domain, and only a few experiments (50 test images) were conducted in physical world.
Another work \cite{thys2019fooling} did not conduct statistical evaluations, and only illustrated a few examples of stealth attack results indoors.

This paper proposes a novel three-dimensional (3D) person stealth attack in real physical world.
First, we design the loss function by targeting the object detector, and generate the adversarial patch through the backpropagation optimization, where several 3D physical constraints are considered and addressed. Then, the adversarial patch is printed on real clothes to make the 3D invisible cloaks.
Anyone wearing the cloak can successfully avoid the detection of object detectors and become ``invisible'' in various indoor and outdoor physical conditions.
The evaluations on multiple generated 3D invisible cloaks show that, the proposed person stealth attack is robust even under those complex and challenging physical conditions, \textit{e.g.}, wrinkled, sheltered, curved, and from different wide angles.

The main contributions of this paper are as follows:

\textbf{(1) 3D person stealth attacks.} In this paper, we propose the concept of 3D invisible cloak, and successfully implement the static and dynamic person stealth attacks in 3D space under various indoor and outdoor physical conditions, rather than the 2D plane.
To the best of the authors' knowledge, we are the first to consider and address the impacts of these challenging 3D physical constraints (radian, wrinkle, occlusion, angle) on the person stealth attacks.
The generated adversarial patches are printed on the non-rigid real clothes to make real 3D invisible cloaks.

\textbf{(2) 3D transformations.} In addition to those conventional physical transformations (such as scale and rotate), this paper proposes four 3D physical transformations to mitigate the influences of these challenging 3D physical constraints on person stealth attacks: the different angles of person detection (Angle), the stereoscopic radian on the invisible cloak caused by person's body shape (Radian), the wrinkles on the cloak caused by person's postures (Wrinkle), and the random regions on invisible cloak sheltered by surrounding objects (Occlusion).
During each iteration of optimization, we perform the conventional and 3D transformations on the adversarial patch, to ensure the robustness of the generated 3D invisible cloak in real world.
Experimental results show that the generated 3D invisible cloaks are effective even under those complicated and challenging physical constraints.

\textbf{(3) The exploration on how to implement the optimal person stealth attacks.} Through the methodical experimental designs, this paper explores how to generate the optimal adversarial patch in terms of: 1) What is the optimal attack success rate that a 3D invisible cloak can achieve?
2) What is the color and shape of the invisible cloak that achieves the optimal performance?
3) How to achieve the optimal person stealth attacks?
Specifically, we explore the optimal attack performance of the generated adversarial patch against the YOLO v2 \cite{redmon2017yolo9000} (You Only Look Once version2) detector.

\textbf{(4) Disappeared 3D invisible cloak.} The adversarial patches generated in the experiments can make humans become ``invisible'', but the patches would be recognized as other objects, \textit{e.g.}, ``teddy bear'' or ``kite''. In this paper, we further attempt to generate 3D invisible cloak which can make a person completely disappear by modifying the objective optimization function. In other words, the humans will not be detected as any objects by YOLO v2.

\textbf{(5) Systematic evaluation framework.} Finally, we introduce a systematic evaluation framework to methodically evaluate the attack performances of these generated patches and 3D invisible cloaks, including the following aspects: 2D, 3D, digital domain, physical world, static and dynamic evaluation, different angles and distances, various indoor/outdoor conditions, and those complicated physical scenarios (such as wrinkle, occlusion, radian, angle).
We have generated massive (over 100) adversarial patches and printed many 3D invisible cloaks.
The attack success rate of the generated adversarial patch in digital domain is 86.56\%, while the attack performances of 3D invisible cloaks in various static and dynamic person stealth attacks is up to 100\% and 77\%, respectively, which is much better than that in existing works.

This paper is organized as follows. Section \ref{sec:sec2} introduces the related works.
Section \ref{sec:sec3} presents the attack model.
Section \ref{sec:sec4} elaborates the proposed person stealth attack method.
Section \ref{sec:sec5} explores how to generate the optimal 3D invisible cloak.
Section \ref{sec:sec6} analyzes the experimental results.
Section \ref{sec:sec7} discusses how to generate a completely disappeared 3D invisible cloak. This paper is concluded in Section \ref{sec:sec8}.

\vspace{-0.25cm}

\section{Related Work}
\label{sec:sec2}

\vspace{-0.2cm}
Recently, generating adversarial examples to fool neural network (NN) based classifiers or object detectors is an active research topic. The inspiration of this paper comes from the adversarial example attacks. Therefore, in this section, we review the adversarial example attacks in digital domain and in real physical world, as well as few co-current works about person stealth attacks.

\vspace{-0.25cm}
\subsection{Adversarial Example Attacks}
\textbf{Digital adversarial example attacks.} Szegedy \textit{et al.} \cite{szegedy2013intriguing} first demonstrate that DNN models are vulnerable to digital adversarial examples. They generate imperceptible perturbations on input images, which successfully cause a DNN based image classifier to output incorrect predictions. Goodfellow \textit{et al.} \cite{goodfellow2014explaining} propose the Fast Gradient Sign Method (FGSM) to accelerate the generation of adversarial examples. Papernot \textit{et al.} \cite{papernot2016limitations} propose the Jacobian-based Saliency Map Attack algorithm (JSMA), which can cause the DNN model to incorrectly classify an adversarial example into a target class. Su \textit{et al.} \cite{su2019one} propose a Differential Evolution (DE) based adversarial example generation method, which only needs to modify one pixel on an image to make the classifier output incorrect results. Moosavi-Dezfooli \textit{et al.} \cite{moosavi2016deepfool} propose the DeepFool to generate universal adversarial perturbations, in which any image added with such perturbations can fool the DNN.

\textbf{Physical adversarial example attacks.} In real world, the input of the DNN model is captured by a camera, thus an attacker cannot modify the model's inputs at the digital level. Recently, constructing physical adversarial examples and launching attacks in real world have attracted more attentions.

Kurakin \textit{et al.} \cite{kurakin2016adversarial} print the digital adversarial examples generated by FGSM algorithm \cite{goodfellow2014explaining}, and then use a camera to take photos for them to generate the physical adversarial examples to fool the image classifier.
Athalye \textit{et al.} \cite{athalye2018synthesizing} propose an Expectation Over Transformation (EOT) algorithm to improve the robustness of the generated adversarial examples in real world.
The EOT algorithm performs a series of transformations on the adversarial example to simulate the physical transformations it may undergo in the real world \cite{athalye2018synthesizing}.
Chen \textit{et al.} \cite{chen2018shapeshifter} propose ``ShapeShifter'' to generate robust physical adversarial examples against the object detector.

Eykholt \textit{et al.} \cite{eykholt2018robust} propose Robust Physical Perturbation (RP2). They generate the digital perturbations and perform transformations to craft the physical stickers.
They use a mask matrix to limit the perturbation region at the target region (such as the road sign) \cite{eykholt2018robust}. Their subsequent work \cite{song2018physical} extends the RP2 algorithm and propose the ``Disappearance Attack'' against the YOLO detectors. By generating the physical stickers and pasting them on the traffic signs, they make the YOLO ignore the existence of these traffic signs \cite{song2018physical}.
Jan \textit{et al.} \cite{jan2019connecting} propose Digital-to-Physical (D2P) method, which uses an image conversion neural network to perform the physical transformation on digital images. They directly add the perturbations into these transformed images to generate physical adversarial examples \cite{jan2019connecting}.
Sharif \textit{et al.} \cite{sharif2016accessorize} propose the physical attacks against the face recognition system. They generate the digital adversarial perturbations by minimizing the cross-entropy loss of the DNN model, and print the perturbations on a pair of glasses. The attacker wearing such glasses can evade the face recognition system \cite{sharif2016accessorize}.

\vspace{-0.25cm}
\subsection{Co-current and Related Work} \label{sec:sec2.B}
To the best of the authors' knowledge, there are only few studies about the person stealth attacks.
Yang \textit{et al.} \cite{yang2018building} construct a pinhole camera model based on the EOT algorithm \cite{athalye2018synthesizing}.
They perform transformations (shift and rotate) on the generated adversarial examples in three directions of the coordinate system of the camera, to simulate the transformations that the adversarial example may undergo in real world \cite{yang2018building}.
However, most of their experiments are carried out in digital domain, while only a few evaluations (50 pictures) are performed in real world.
Moreover, the physical stealth attacks are evaluated by displaying the generated adversarial stickers on the screen of a tablet, thus the work \cite{yang2018building} does not consider the color distortion problem during the printing process. Besides, they didn't consider the complex 3D challenging conditions.
Finally, they lack the evaluation of outdoor physical conditions and dynamic physical stealth attacks.

The work in parallel with this paper is Thys \textit{et al.} \cite{thys2019fooling}, they generate the adversarial patch by minimizing the detection score of YOLO v2 detector for the class ``person''. They perform several conventional transformations (rotate, scale, brightness, random noise) on a generated patch to improve its robustness.
However, the work \cite{thys2019fooling} doesn't consider those 3D physical constraints during the realistic person detection process, and only prints the patch on a $40cm*40cm $ cardboard, which means their proposed stealth attack still remains on the 2D plane. In addition, they lack statistical evaluations of stealth attack in digital domain and real physical world. Only a few examples of stealth attack results are illustrated in \cite{thys2019fooling}.

Compared with the existing person stealth attacks \cite{yang2018building, thys2019fooling}, the advantages and differences of this paper are as follows:
\begin{itemize}
\vspace{-0.25cm}
  \item {\textbf{3D physical transformations.} The proposed person stealth attack considers challenging 3D physical constraints (Radian, Wrinkle, Angle, Occlusion) and propose 3D wearable physical transformations when generating the adversarial patches. In this way, even the generated invisible cloak has under various complex physical conditions (\textit{e.g.}, wrinkled, sheltered, curved, and from different angles), it is still effective for person stealth attacks in real world.}
      \vspace{-0.25cm}
  \item {\textbf{3D targeted attack.} The proposed person stealth attack focus on practical 3D physical attacks instead of 2D plane or digital simulation. Except for the conventional physical constraints, this paper mitigates the influences of those 3D physical constraints on the person stealth attacks. Besides, we print the adversarial patches on real clothes, rather than simply displayed on a tablet computer \cite{yang2018building} or printed on a rigid cardboard \cite{thys2019fooling}.}
      \vspace{-0.25cm}
  \item {\textbf{Exploration on how to implement the optimal person stealth attacks.} The existing person stealth attacks \cite{yang2018building, thys2019fooling} have not discussed the impacts of original images on the performances of generated adversarial patches. This makes their person stealth attacks not the optimal.
      In this paper, through the systematic experimental explorations and demonstrations, we explore: 1) What is the optimal attack success rate that a 3D invisible cloak can achieve? 2) What is the color and shape of the invisible cloak that achieves the optimal performance? 3) How to achieve the optimal person stealth attacks? In this way, the proposed attack method can achieve the optimal performance against the object detector.}
\vspace{-0.25cm}
  \item {\textbf{Systematic evaluations.} The existing studies \cite{yang2018building, thys2019fooling} lacked sufficient evaluations for their proposed attack methods, and most of their evaluations are performed in digital simulation or only remain on 2D physical plane. This paper presents a systematic and comprehensive evaluation framework to methodically evaluate the proposed person stealth attacks, including 2D, 3D, digital domain, physical world, static and dynamic evaluation, different angles and distances, various
indoor/outdoor conditions, and those complicated physical scenarios (wrinkled, sheltered, curved, and from different angles), \textit{etc}.}
\end{itemize}

\vspace{-0.25cm}

\section{Attack Model, Attackers' Challenges and Solutions}
\label{sec:sec3}
\vspace{-0.25cm}
In this section, we formulate the attack model from four aspects: the attacker's goal, knowledge, capability, and strategy.
Besides, we analyze the technical challenges about launching the physical person stealth attacks, and present the solutions.

\vspace{-0.25cm}
\subsection{Attack Model}
\vspace{-0.25cm}

\textbf{Attacker's goals.} In this paper, the attacker's goal is to achieve stealth by wearing the 3D invisible cloak to evade the detection of YOLO v2 \cite{redmon2017yolo9000} detector in various indoor and outdoor scenes.
Note that, the overall procedure of the proposed person stealth attack is universal.
By slightly modifying the internal parameters (\textit{e.g.}, the network configurations and training weights) during the backpropagation optimization process, the attacker can generate the adversarial patches and craft the 3D invisible cloaks against other types of object detectors, such as SSD, Faster-RCNN, \textit{etc}.

\textbf{Attacker's knowledge.} We assume that the attacker can attack the person detectors in the white-box setting, \textit{i.e.}, the attacker has the following knowledge: the type of the detector that used in the system; the working principle of the object detector. There are two reasons for making such assumptions on the attacker's knowledge.
First, in order to develop a secure and effective defense technique, it is necessary to evaluate the security performance of a system when facing strong attackers. Second, in practice, it is easy for an attacker to obtain the above information of a deployed system.

\textbf{Attacker's capabilities.} In digital simulation experiments \cite{carlini2017towards,szegedy2013intriguing,goodfellow2014explaining,papernot2016limitations,su2019one,moosavi2016deepfool}, the attacker can directly add digital perturbations in the input of a DNN model to fool the model. However, for an object detector that deployed in real world, the input of the DNN model is captured by a camera, thus the attacker cannot directly modify the input at the digital level. Therefore, in this paper, the attacker can only perform the person stealth attack against the YOLO v2 detector by generating robust physical adversarial examples \cite{kurakin2016adversarial,athalye2018synthesizing,eykholt2018robust,jan2019connecting} in real world.

\textbf{Attacker's strategy.}
The attacker's strategy is to construct an adversarial patch that remains effective under difficult realistic physical constraints.
In other words, this patch can fool the YOLO v2 \cite{redmon2017yolo9000} object detector to ignore the presence of a ``person''.
In this paper, the attacker first constructs a rectangular adversarial patch with the size of $29cm * 43cm$ in digital domain, and then performs a series of physical transformations on it.
By printing the adversarial patch on real clothes to make the 3D invisible cloak, the attacker can wear such clothes to evade the detection of object detector.

\vspace{-0.25cm}
\subsection{Attackers' Challenges}
\label{sec:sec4.1.2}
The attackers face a range of challenges when launching the person stealth attacks in real world.
The main challenge is that, the generated adversarial patches need to be robust against various physical constraints, such as the changes in lighting conditions and angles.
Those physical constraints will result in the perturbations added in the patches to be ineffective in real world, thus cause the physical stealth attacks to fail.

\textbf{Conventional physical constraints \cite{thys2019fooling}.} Conventional physical constraints include: 1) the size of the object in a captured image varies due to different distances from the detector; 2) the intensity of the light can cause the differences in the brightness of the photos; 3) random noises can interfere with the adversarial patch in the photos.

\textbf{3D physical constraints.} In addition to the above conventional physical constraints, we consider the 3D physical constraints for the first time, and make the 3D invisible cloaks. These 3D physical constraints are as follows:
\begin{enumerate}
  \item {Radian: people with different shapes will produce stereoscopic radians in different areas of the invisible cloak.}
  \item {Wrinkle: human's postures change dynamically will cause different degrees of wrinkles on the cloak.}
  \item {Angle: different angles the invisible cloak facing the detector will cause the camera to capture different images.}
  \item {Occlusion: the complex physical scenes will cause the patch printed on the invisible cloak to be covered.}
\end{enumerate}

Those considered four physical constraints are the ones that a patch mainly undergoes in real world, which can greatly affect the effectiveness of a generated adversarial cloak.
Taking the constraint ``Radian'' as an example, the attackers in real world may have different shapes of bodies.
As a result, different areas of a cloak will produce different degrees of stereoscopic radians, which depend on the attacker's body shape.
Such stereoscopic radians on the invisible cloak will degrade the attack performance of the generated 3D invisible cloak in physical world.

\vspace{-0.25cm}
\subsection{Solutions}

To solve the above challenges, we perform a series of physical transformations on the adversarial patch during each iterative optimization.
These transformations are based on the \textit{Expectation Over Transformation (EOT)} attack framework \cite{athalye2018synthesizing}.
However, unlike \cite{athalye2018synthesizing} which generates the physical adversarial example targeting the image classifiers, this paper launches the person stealth attacks targeting the DNN based object detectors which are more challenging.
Moreover, for the first time, we focus on addressing the impacts of those 3D challenging physical constraints on the performance of an invisible cloak.

The physical transformations considered in this work include: 1) the conventional transformations $R$; 2) 3D transformations $U$.
The conventional transformations $R$ is proposed by Thys \textit{et al.} \cite{thys2019fooling}, including scaling the size of the patch randomly, rotating the patch with random angles in the plane, adjusting the brightness of the patch, and adding random noises into the patch. We propose the 3D transformations $U$, which includes the overlay of the wrinkle patterns, the overlay of the radian patterns, the forward and backward rotation of the generated patches at random angles, and the occlusion of random regions in the patch.
\vspace{-0.25cm}

\section{The 3D Invisible Cloak Generation Method}
\label{sec:sec4}
\vspace{-0.2cm}

In this section, first, the overall procedure of the proposed stealth attack is described. Then, the 3D invisible cloak generation method for stealth attacks is elaborated. Finally, the proposed evaluation framework is presented.

\vspace{-0.25cm}
\subsection{Overall Procedure of the Proposed Stealth Attack} \label{overall}
\begin{figure}[htbp]
  \centering
  \includegraphics[width=3.1in]{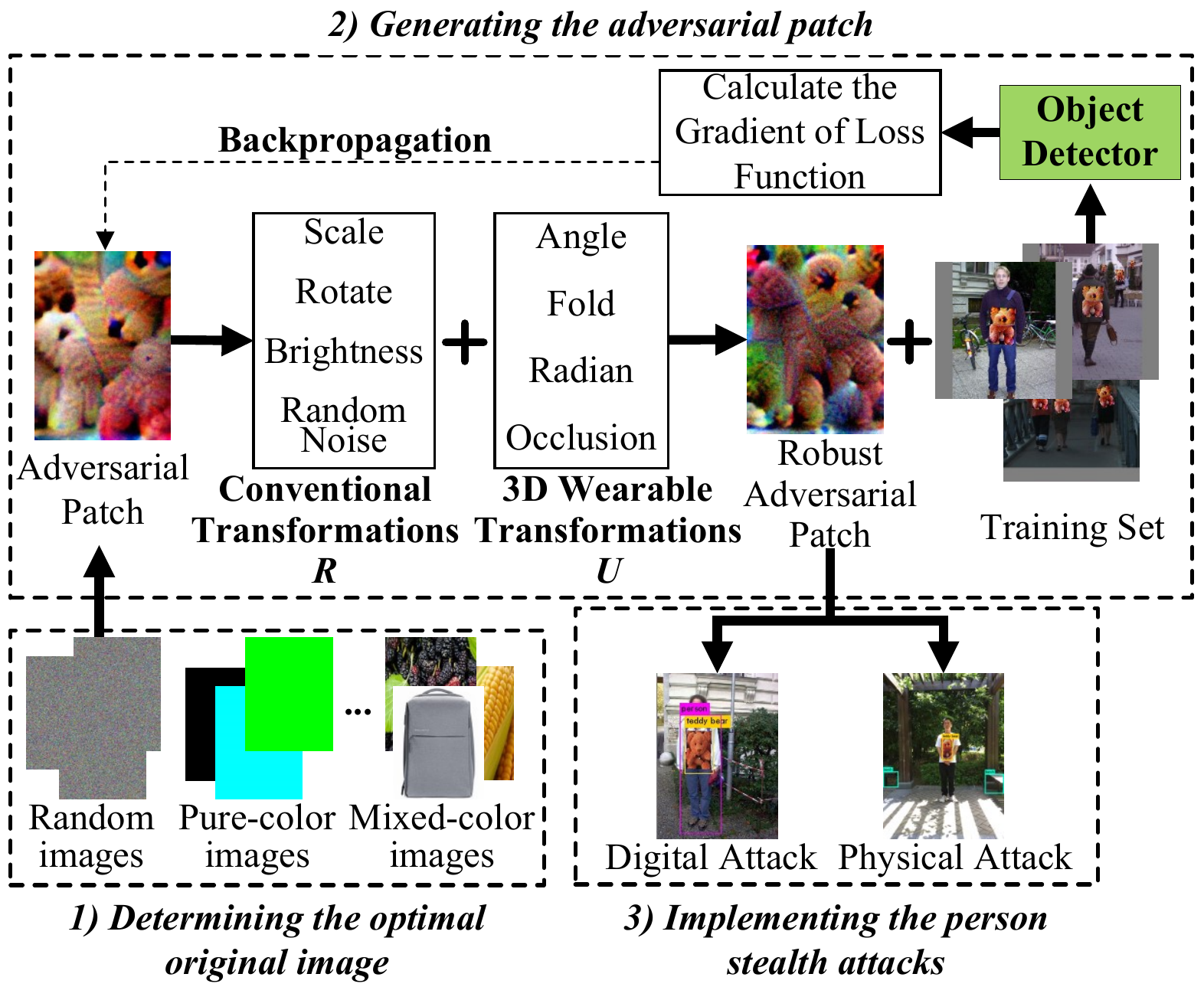}\\
  \caption{Overall procedure of the proposed person stealth attack.}
  \label{fig:fig1}
\end{figure}

The proposed stealth attack against the person detectors can be divided into following steps.
First, determining the optimal original image. This steps aims to find the input image which can generate the adversarial patch with the optimal stealth attack performance. The process of selecting the original images are discussed in Section \ref{sec:sec5}.
Second, generating the robust adversarial patch from the selected image through the iterative optimization.
After the above two steps, the proposed attack method can determine the optimal original image and generate the optimal adversarial patch.
Finally, using this adversarial patch and its generated 3D cloak to implement the person stealth attack in digital domain and in real-world conditions, respectively.
Figure \ref{fig:fig1} shows the overall procedure of the proposed person stealth attack.

\textbf{Determining the optimal original image.} An adversarial patch is a set of adversarial perturbations, which are generated from some original pixel values through the backpropagation optimization process.
For the proposed stealth attack, these original pixel values can come from any image, such as an image that entirely composed of random pixel values, or a real image that contains some objects.
In this paper, we explore how to select the original image to generate the adversarial patch which can achieve the optimal attack performance. The process of image selections are discussed in Section \ref{sec:sec5}.
In this way, we can determine which type (shapes, colors, etc.) of the original image can generate the optimal adversarial patch.

\textbf{Generating the adversarial patch.} First, a clean input image is used as the initial adversarial patch ${P^*}$.
Then, a series of physical transformations (including the conventional transformations $R$ and 3D transformations $U$) are performed on this patch to enhance its robustness in real physical world.
Third, the transformed patch ${P^*}$ is pasted on the images of Inria training set \cite{dalal2005histograms} that contain humans, and then it will be submitted to target object detector for person detection.
Finally, the gradient of the objective loss function is computed according to the detection result.
Specifically, the gradient descent based optimization method, Adam algorithm \cite{kingma2015adam}, is used to backpropagate on the network and iteratively optimizes the adversarial patch ${P^*}$ until the value of the objective loss function is less than a predefined threshold.

\textbf{Implementing the person stealth attacks.} Once the robust adversarial patch ${P^*}$ has been generated, it will be used to launch the person stealth attack. In digital domain, the patch ${P^*}$ will be pasted on the person in the images of Inria test set \cite{dalal2005histograms} to implement the stealth attacks. In real physical world, the patch ${P^*}$ will be printed on real clothes to generate the 3D invisible cloaks to conduct the static (taking photos) and dynamic (recoding videos) person stealth attacks.

\vspace{-0.3cm}
\subsection{3D Invisible Cloak Generation Method} \label{generation_method}
In order to prevent YOLO v2 \cite{redmon2017yolo9000} detector from detecting the presence of the ``person'', the attacker must ensure that the added adversarial patch can make the confidence of the class ``person'' in the prediction as low as possible. In this way, since the prediction confidence of the class ``person'' is lower than the threshold of the system, YOLO v2 will ignore the existence of ``person'' in its input. In this paper, the objective loss function of person stealth attack is defined as follows \cite{song2018physical}:
\begin{equation}
  {L_F}({x_{in}},{P^ * },{y_{per}}) = \mathop {\max }\limits_{\scriptstyle s \in S*S \hfill \atop
  \scriptstyle b \in B \hfill}  {E_p}(s,b,{F_\rho }({x_{in}} + {P^ * }), {y_{per}})
\end{equation}
where ${x_{in}}$ is the clean input image that contains the person, ${P^*}$ is the generated adversarial patch, and ${y_{per}}$ represents the true label corresponding to the person \cite{song2018physical}. $S$ is the number of square grid cells of the input image ${x_{in}}$. $B$ is the number of bounding boxes contained in each grid cell. ${F_\rho }$ is the output function of YOLO v2 detector, while ${E_p}$ is the function that extracts the prediction confidence of the person from the bounding boxes generated in image ${x_{in}}$ \cite{song2018physical}. The loss function ${L_F}$ will output the maximum confidence score of the person ${y_{per}}$ in image ${x_{in}}$. Finally, the gradient descent based Adam algorithm \cite{kingma2015adam} is used to optimize the objective loss function, until the adversarial patch can successfully make YOLO v2 ignore the presence of the person ${y_{per}}$.

In this work, the generated adversarial patch needs to be printed on real clothes to generate the 3D invisible cloak.
To avoid alertness caused by intense visual conflict after wearing the invisible cloak, the generated adversarial patch should look natural. In other words, the pixel value of any point on the patch should be as close as possible to the pixel value of its surrounding points. Therefore, we add the total variation (TV) constraint which was proposed in \cite{mahendran2015understanding} as the non-smoothness penalty term of the generated patch \cite{mahendran2015understanding}:
\begin{equation}
  TV({P^ * }) = {\sum\limits_{i,j} {({{({P_{i,j}} - {P_{i + 1,j}})}^2} + {{({P_{i,j}} - {P_{i,j + 1}})}^2})} ^{{\raise0.5ex\hbox{$\scriptstyle 1$}
\kern-0.1em/\kern-0.15em
\lower0.25ex\hbox{$\scriptstyle 2$}}}}
\end{equation}
where ${P_{i,j}}$ represents the pixel value of the point whose coordinate position is $(i,j)$ on the patch ${P^ * }$.

In addition, even the most advanced color printers cannot fully reproduce all the colors in an digital image, which will result in color distortion in the adversarial patch ${P^ * }$.
For this reason, we add the constraint item of non-printable score (NPS) \cite{sharif2016accessorize} to the objective optimization function to ensure the printability of the generated adversarial patch \cite{sharif2016accessorize}:
\begin{equation}
NPS(P^{*})=\prod\limits_{\scriptstyle p^{'}\in P^{*} \hfill \atop
  \scriptstyle p^{''}\in P_{color} \hfill} \left |p^{'}- p^{''}\right |
\end{equation}
where $P^{*}$ is the generated adversarial patch and ${P_{color}} \subset {[0,1]^3}$ is a set of printable RGB colors \cite{sharif2016accessorize}. A lower NPS score means that the adversarial patch printed in real physical world is closer to its digital version.

In summary, in order to ensure the robustness of the generated adversarial patch in real physical world, the final objective optimization function to generate 3D wearable invisible cloak is as follows:
\begin{equation} \label{eq:eq5}
\mathop {\arg \min }\limits_{{P^ * }} {\kern 1pt} {E_{\scriptstyle {t_r} \sim R \hfill \atop
  \scriptstyle {t_u} \sim U \hfill} }[{L_F}({x_{in}},{t_u}({t_r}({P^ * })),{y_{per}}) + {\lambda _1}TV + {\lambda _2}NPS]
\end{equation}
where $R$ is the conventional transformations, including patch rotation, scaling, brightness adjustment, and adding random noise. $U$ represents the 3D physical transformations, including the superposition of radian and wrinkle, the transformation on patch's angle, and the object occlusion of random regions in the patch. The parameters ${\lambda _1}$ and ${\lambda _2}$ are used to adjust the weight of the penalty term $TV$ and $NPS$.

\vspace{-0.25cm}
\subsection{Evaluation Framework}
\begin{figure}[htbp]
  \centering
  \includegraphics[width=2.4in]{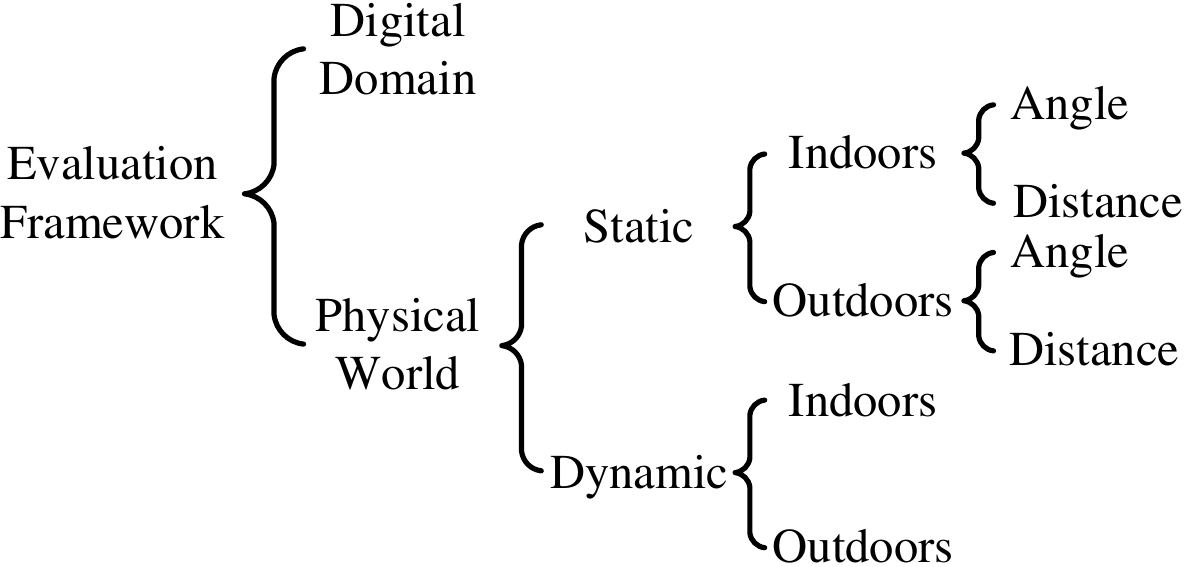}\\
  \caption{The proposed evaluation framework for person stealth attacks.}
  \label{fig:fig2}
\end{figure}

The existing stealth attack works \cite{yang2018building,thys2019fooling} lack comprehensive and systematic evaluations. Most of the evaluations in \cite{yang2018building} are digital simulation experiments, while a few experiments in real world are only the static evaluations indoors, which remain on the 2D plane.
Although static and dynamic stealth attack experiments have been performed in \cite{thys2019fooling}, their evaluations in digital domain and real physical world lack statistical experimental results, and also remain on the 2D plane.
This paper proposes a systematic evaluation framework for person stealth attacks, as shown in Figure \ref{fig:fig2}.

\textbf{Digital domain.} The evaluation framework first evaluates the attack performance of the stealth attacks in digital domain. This is done by directly pasting the generated adversarial patch onto the person in the images of Inria test set \cite{dalal2005histograms}.

\textbf{Physical world.} The evaluations in physical world are to print the generated patches on real clothes to make the 3D invisible cloaks. The evaluation includes the static method and the dynamic method. The static evaluation is to let the testers wear the cloaks and take photos for them at various angles and distances under indoor and outdoor scenarios, and then submit these photos to YOLO v2 for detection. The dynamic evaluation is to evaluate the effectiveness of 3D invisible cloak in real-time detection scenarios, where the testers wearing the cloaks are moving and videos are recorded.

\vspace{-0.25cm}
\section{Optimal Adversarial Patch}
\label{sec:sec5}
\vspace{-0.25cm}
In this section, we will explore how to generate the 3D invisible cloak that can achieve the optimal performance of person stealth attacks.

\textbf{Questions.}
In theory, any image can be used to generate the adversarial patch, but different original images obviously have different impacts on the performance of the generated adversarial patch. Related stealth attack works \cite{yang2018building,thys2019fooling} lack necessary discussions about the original image that used to generate the adversarial patch, which cause the stealth attack performances of their adversarial patches are different and may not be optimal. In this paper, we have the following questions about the generated adversarial patch:
\begin{itemize}
\vspace{-0.25cm}
  \item {\textbf{Q1}) What is the highest attack performance of a generated adversarial patch in the physical world?}
      \vspace{-0.25cm}
  \item {\textbf{Q2}) What shape does the adversarial patch with the optimal attack performance visually look like, and what is the result detected by the object detector?}
      \vspace{-0.25cm}
  \item {\textbf{Q3}) How to generate the adversarial patch with the optimal attack performance, and what conditions should the original image meet?
}
\end{itemize}

\textbf{Exploration processes of generating optimal adversarial patches.} Using a range of exploratory experiments, we find that there are two factors that affect the stealth attack performance of the adversarial patch: the color of the original image ($Color$), and the shape ($Shape$) of the objects contained in the image.
To address these questions, we use methodological experimental procedures to explore the answers, and report the experimental results in Section \ref{sec:sec6}. The exploration process in this paper is divided into the following three steps:
\begin{enumerate}
\vspace{-0.25cm}
  \item {First, we generate the adversarial patches from images that completely composed of random pixel values. In this way, by observing the attack performances of the adversarial patches generated from random images, we can determine whether the selection of an original image is necessary (Section \ref{sec:sec6.2.1});}
      \vspace{-0.25cm}
  \item {Second, we select multiple pure-color images (contain no objects) and mixed-color images (contain some objects) to generate the adversarial patches, respectively. By comparing the performances of the adversarial patches that generated from those pure-color and mixed-color images, we can determine whether the color of an original image and the presence of objects in an image affect the attack performance of the generated adversarial patchs (Section \ref{sec:sec6.2.2});}
      \vspace{-0.25cm}
  \item{Third, we select the images of different colors that contain the same object, and the images of same color that contain different objects to generate the adversarial patches. The former type of images are used to determine which color ${c_k} \in Color$ of the original image is optimal, while the latter type of images are used to determine which shape ${s_k} \in Shape$ of objects contained in the original image is optimal (Section \ref{sec:sec6.2.4}).
  }
\end{enumerate}
Finally, the adversarial patch generated by the original image $<{c_k},{s_k}>$ will reach the highest stealth attack performance (Section \ref{sec:sec6.2.4}).

\textbf{Conclusions.} According to the above experimental explorations, we draw the following important conclusions:
\begin{enumerate}
      \vspace{-0.25cm}
  \item {First, for a specific object detector, the highest success rate of the stealth attack that the generated adversarial patches can achieve is fixed. Taking the YOLO v2 \cite{redmon2017yolo9000} object detector as an example, the stealth attack success rate of the generated adversarial patch keeps around 86.56\% on Inria dataset and 100\% in real physical world.}
      \vspace{-0.25cm}
  \item {Second, the adversarial patch which reaches the optimal performance is generated from the original image whose color is orange and the contained object's shape is a teddy bear.
      The generated patch looks like teddy bears, and will be detected as ``teddy bear'' by YOLO v2.}
      \vspace{-0.25cm}
  \item {Third, any adversarial patch with good stealth attack performance will be detected as ``teddy bear'' by YOLO v2. This also explains why these generated adversarial patches (in this paper and the work \cite{thys2019fooling}) against the YOLO v2 detector are similar in shape, i.e., teddy bear. In fact, they are all working towards the adversarial patch with the optimal performance.}
            \vspace{-0.25cm}
\end{enumerate}

 \vspace{-0.25cm}
\section{Experimental Evaluations}
\label{sec:sec6}
\vspace{-0.25cm}
In this section, we evaluate the stealth attack performance of the adversarial patch and its generated 3D invisible cloak.
The experimental setup is introduced in Section \ref{sec:sec6.1}, which includes the experimental dataset, the target object detector, and the evaluation metric.
First, the stealth attack performances of the adversarial patches generated by random images, different color (pure-color and mixed-color) images, and the optimal adversarial patches in digital domain are presented in Section \ref{sec:sec6.2}. These are used to explore the optimal performance of person stealth attack and its generation method.
Then, the static and dynamic person stealth attack results of the generated 3D invisible cloak in real physical world is presented in Section \ref{sec:sec6.3}.
Lastly, the comparison with existing stealth attack works is presented in Section \ref{sec:sec6.4}.

 \vspace{-0.25cm}
\subsection{Experimental Setup}
\label{sec:sec6.1}
The experimental setup, including the dataset (Inria dataset \cite{dalal2005histograms}), the object detector (YOLO v2 \cite{redmon2017yolo9000}) and the evaluation metric (attack success rate $R_{s}$), are introduced in Appendix \ref{Appendix_F1}.

 \vspace{-0.25cm}
\subsection{Digital Stealth Attacks}
\label{sec:sec6.2}
The person stealth attacks in digital domain are conducted on Inria test set \cite{dalal2005histograms}. In our experiments, in the case of no stealth attacks, a total of 602 people can be detected by YOLO v2 in 288 positive samples of Inria test set. Therefore, we evaluate the performance of person stealth attacks in digital domain by pasting the generated adversarial patch on these 602 people. Since the digital adversarial patch will undergo rotating and scaling before being pasted on the people in the Inria test images, the size and shape of the patch that pasted on the person are different at each time.
As a result, the success rate of person stealth attacks in digital domain varies each time.
Therefore, we repeat the stealth attack 10 times for each generated patch and take the average of them as the final stealth attack success rate of that patch in digital domain.

\vspace{-0.25cm}
\subsubsection{Patches Generated from Random Images} \label{sec:sec6.2.1}
\vspace{-0.25cm}
As discussed in Section \ref{overall}, an adversarial patch can be generated from any original image.
In this paper, first, we evaluate the attack performances of the adversarial patches that generated from these random images (\textit{i.e.}, completely filled with random pixel values), so as to analyze whether the selection of original images is necessary.
Specifically, we construct 10 different random images, and generate an adversarial patch for each image with the proposed stealth attack method.
These generated patches are pasted on the people in the Inria test set to evaluate their stealth attack effectiveness.
The experimental results show that, their average stealth attack success rate is 54.27\%, and their maximum and minimum attack success rate are 56.01\% and 52.60\%, respectively.
This indicates the adversarial patches generated from random images are effective for person stealth attacks, but obviously they are not the optimal.
In order to achieve the optimal performance of person stealth attack, it is necessary to generate the adversarial patches from selected original images, rather than those random ones.

 \vspace{-0.25cm}
\subsubsection{Patches Generated from Different Color Images}
\label{sec:sec6.2.2}
 \vspace{-0.25cm}
In this section, we evaluate the effectiveness of adversarial patches that generated from the images of different colors.

\textbf{Pure-color images.} First, we select 16 representative colors: \{red, orange, yellow, green, blue, purple, black, white, gray, cyan, magenta, pink, olive, navy, wine, brown\}, as the filling colors of the initial images to generate the adversarial patches, respectively.
The above chosen colors are comprehensive and include various types: 1) primary color; 2) basic color; 3) intermediate color; 4) other popular colors, which cover almost all the commonly used colors in the daily life.

\textbf{Mixed-color images.} A mixed-color image has a main color, and also contains some objects with other colors.
Figure \ref{fig:fig4} in Appendix \ref{Appendix_F3} shows some mixed-color images that dominated by yellow, green, and blue, as well as their generated adversarial patches.
In our experiments, we select 10 most commonly used colors form those pure colors: $Color$ = \{red, orange, yellow, green, blue, purple, brown, black, white, gray\}, as the main colors of these mixed-color images to generate the adversarial patches, respectively.

We present the adversarial patches generated from the pure-color and mixed-color images, and analyze their stealth attack performances in digital domain in Appendix
\ref{Appendix_F3}. The experimental results on those pure-color and mixed-color images demonstrate that, both the color of the original image and the presence of objects in an image have great impacts on the performance of the generated adversarial patch.

\vspace{-0.3cm}
\subsubsection{Optimal Adversarial Patches}
\label{sec:sec6.2.4}
\vspace{-0.25cm}
We explore which color of input image and what shape of objects it contained can generate the optimal adversarial patch.

First, we randomly select two original images that contains different objects (banana and candy), and generate different adversarial patches by changing their colors. The selected colors are the three best colors from each color level (as discussed in Appendix \ref{Appendix_F3}): orange, purple, and green. Their stealth attack results on Inria test set are shown in Table \ref{tab:tab3}. The results further demonstrate that the stealth attack success rates of the adversarial patches indeed vary depending on the color of the original images. Besides, the success rate of stealth attacks corresponds to the level of the original image's color, \textit{i.e.}, orange is better than purple, and purple is better than green (as discussed in Appendix \ref{Appendix_F3}).

\vspace{-0.15cm}
\begin{table}[htbp]
\renewcommand{\arraystretch}{0.7}
\linespread{0.7}
  \centering
  \footnotesize
  \caption{The attack results of adversarial patches generated from original images of different colors}
  \setlength{\tabcolsep}{0mm}{
    \renewcommand\arraystretch{-1}
    \begin{tabular}{m{4.2em}<{\centering}m{4em}<{\centering}m{6.5em}<{\centering}m{4.2em}<{\centering}m{4em}<{\centering}m{6.5em}<{\centering}}
    \toprule
    \textbf{Original image}    & \textbf{Image color}    & \textbf{Attack success rate $R_{s}$}    & \textbf{Original image}    & \textbf{Image color}    & \textbf{Attack success rate $R_{s}$} \\
    \midrule
    \multirow{3}[13]{*}{Banana} & Orange     & 83.38\% & \multirow{3}[13]{*}{Candy} & Orange & 75.59\% \\
\cmidrule{2-3}\cmidrule{5-6}         & Purple & 75.91\% &        &Purple & 71.69\%\\
\cmidrule{2-3}\cmidrule{5-6}         & Green  & 60.96\% &        &Green  & 61.04\% \\
    \bottomrule
    \end{tabular}%
    }
  \label{tab:tab3}%
\end{table}%

Then, we select multiple images of the optimal color (orange) but contain different objects to generate the adversarial patches.
Note that, YOLO v2 detector can recognize 80 different classes of objects at most.
In our experiments, we randomly select 21 classes (``teddy bear'' and 20 other classes) from those 80 classes.
For each class, we use an original image that contains the objects of this class to generate one adversarial patch.
The objects in these original images can be correctly recognized by YOLO v2 before the adversarial transformations.
The reason why we pay special attention to the class of ``teddy bear'' is that, according to the experimental results in Appendix \ref{Appendix_F3}, the stealth attack success rate of adversarial patches which visually look like ``teddy bear'' are relatively high.
The experimental results of these 21 generated adversarial patches on Inria test set are shown in Table \ref{tab:tab4}.
It is shown that, the adversarial patch generated from the original image which contains ``teddy bear'' has the highest stealth attack success rate (84.81\%).
The average stealth attack success rate of these adversarial patches generated from other 20 classes is 71.70\%, and their maximum attack success rate is 82.76\%.
This is because, compared to the original images that contain other shapes of objects, the original image with the ``teddy bear'' pattern is more likely to be optimized to the adversarial patch with the shape of ``teddy bear'', thus can achieve higher stealth attack success rate.

\begin{table}[htbp]
\renewcommand{\arraystretch}{0.7}
\linespread{0.8}
  \centering
  \footnotesize
  \caption{The attack results of adversarial patches generated from original images of the same color but contain different objects}
    \begin{tabular}{cccc}
    \toprule
    \multirow{2}[2]{*}{\textbf{\tabincell{c}{Main color \\of images}}} & \multirow{2}[2]{*}{\textbf{\tabincell{c}{Objects in \\images}}} & \multicolumn{2}{c}{\multirow{2}[2]{*}{\textbf{\tabincell{c}{Attack success \\rate $R_{s}$}}}} \\
          &       & \multicolumn{2}{c}{} \\
    \midrule
    \multirow{4}[10]{*}{Orange} & Teddy bear & \multicolumn{2}{c}{84.81\%} \\
\cmidrule{2-4}          & \multirow{3}[6]{*}{\tabincell{c}{Other \\20 classes}} & Average & 71.70\% \\
\cmidrule{3-4}          &       & Maximum & 82.76\% \\
\cmidrule{3-4}          &       & Minimum & 42.90\% \\
    \bottomrule
    \end{tabular}%
  \label{tab:tab4}%
\end{table}%

\begin{figure}[htbp]
  \centering
  \setlength{\belowcaptionskip}{-0.25cm}
  \includegraphics[width=1.55in]{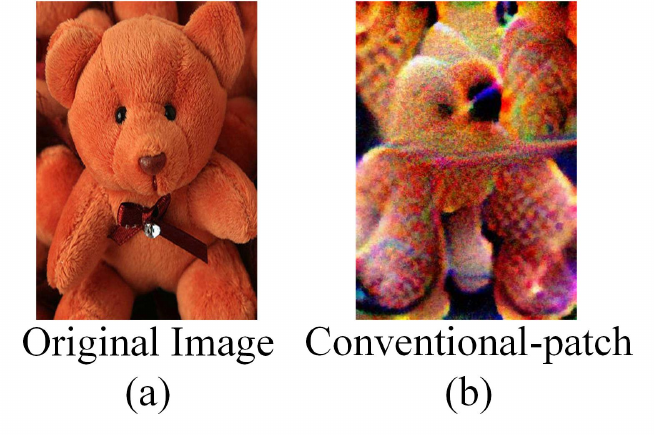}\\
  \caption{The original image and its generated Conventional-patch.}\label{fig:fig5}
\end{figure}

Based on the above discussions, we use the original image with the main color of ``Orange'' and the shape of ``teddy bear'' to generate the final adversarial patch, and the generated patch can achieve the optimal attack performance.
The original image and its generated optimal adversarial patch under conventional transformations are shown in Figure \ref{fig:fig5}. Figure \ref{fig:fig5}(a) shows the original image (an orange teddy bear) that used to generate the adversarial patch, and Figure \ref{fig:fig5}(b) shows the final generated patch. For the convenience of description, these patches and their generated invisible cloaks are named using the physical constraints they have been considered. For example, the adversarial patch shown in Figure \ref{fig:fig5}(b) will be referred to as ``Conventional-patch'', and the generated invisible cloak will be referred to as ``Conventional-cloak''.

\begin{figure}[H]
  \centering
  \includegraphics[width=3.0in]{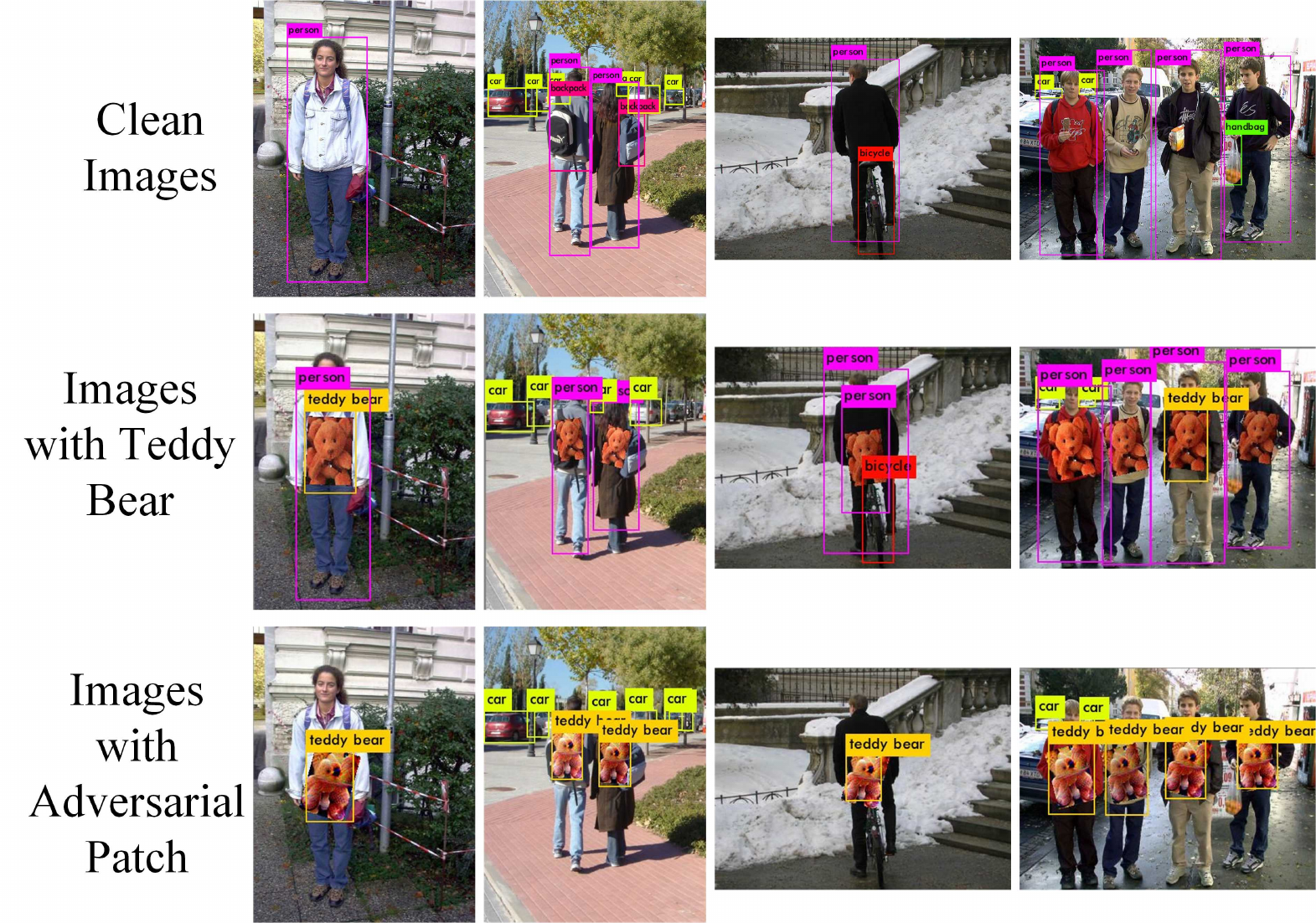}\\
  \caption{Some stealth attack results on Inria test images using the original teddy bear image and its generated adversarial patch.}\label{fig:fig6}
\end{figure}

For comparison, we separately paste the original teddy bear image and the generated adversarial patch on the person in Inria test images, to launch the person stealth attack in digital domain.
The results show that, the stealth attack success rate of original image and adversarial patch is 9.38\% and 84.81\%, respectively.
This means the original image without adversarial transformations is ineffective for person stealth attacks, while its generated adversarial patch can achieve significant attack performance.
Figure \ref{fig:fig6} shows some examples of person stealth attacks conducted on Inria test images using the original teddy bear image and its adversarial patch. The first row are the unprocessed Inria test images, and the second row are the images that pasted with the original teddy bear.
In the third row, the adversarial patch (Figure \ref{fig:fig5}(b)) is pasted on the front or back of the person in the images.
The person pasted with the original teddy bear image can still be detected, while the person pasted with its generated adversarial patch can evade the detection of YOLO v2 and achieve stealth.

 \vspace{-0.25cm}

\vspace{-0.25cm}
\subsection{Physical Stealth Attacks}
\label{sec:sec6.3}
\begin{figure}[htbp]
  \centering
  \includegraphics[width=3.05in]{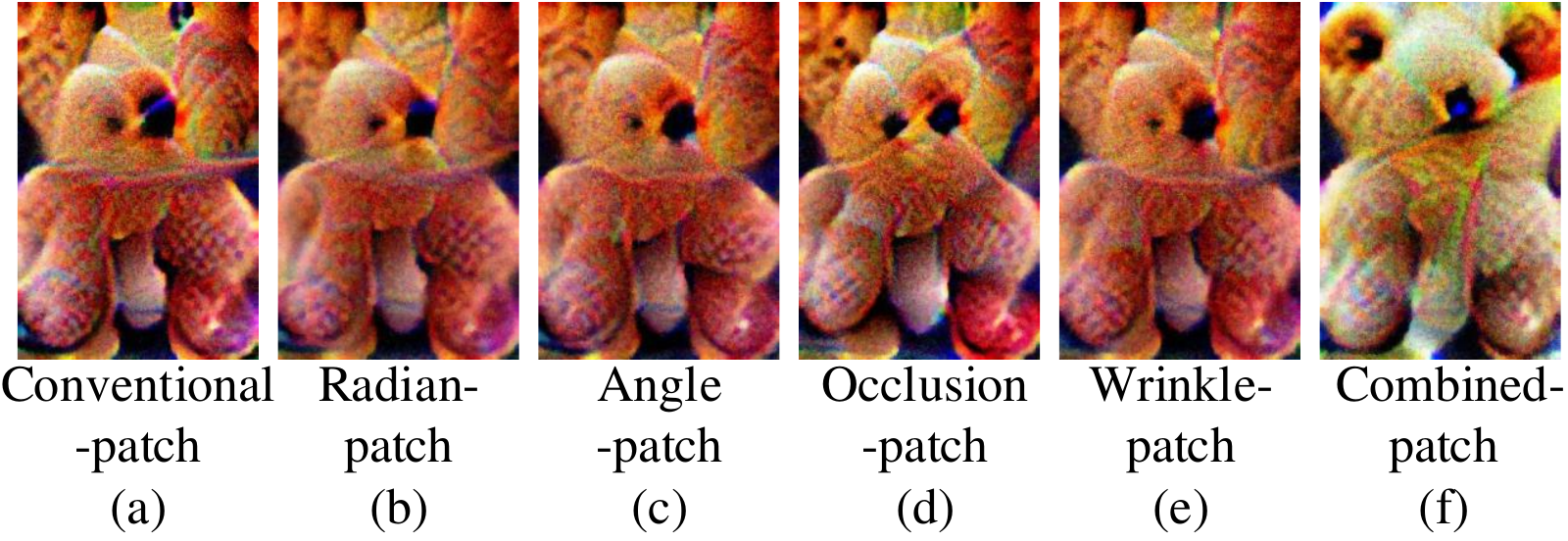}\\
  \caption{Six adversarial patches with different physical transformations for physical stealth attacks.}\label{fig:fig7}
\end{figure}

 \vspace{-0.25cm}
To evaluate the influences of different physical constraints on the stealth attacks, we generate various adversarial patches as shown in Figure \ref{fig:fig7}, which includes: 1) the patch generated under the conventional transformations (scale, rotate, light, and random noise), as shown in Figure \ref{fig:fig7}(a); 2) the patches generated by additionally considering the 3D physical constrains (\textit{i.e.}, radian, wrinkle, angle and occlusion), as shown in Figure \ref{fig:fig7}(b)$\sim$\ref{fig:fig7}(e), respectively; 3) the patch generated by considering all the above constraints is shown in Figure \ref{fig:fig7}(f).
Figure \ref{fig:fig8} in Appendix \ref{Appendix9.4} shows the generated 3D invisible cloaks that printed with different adversarial patches.

 \vspace{-0.25cm}
\subsubsection{Static Physical Stealth Attacks} \label{static_attack}
 \vspace{-0.25cm}
For the static evaluations, we take photos for the testers who wear the invisible cloaks at various distances and angles under the indoors and outdoors physical conditions, and then submit these photos to YOLO v2 for person detection.
Specifically, the test photos are captured from different people in different places under various weather conditions.
In this paper, the outdoor photos are mainly captured from the following scenes: 1) the aisle under the overpass (sunny, morning); 2) the school playground (cloudy, afternoon); 3) the parking lot (cloudy, evening).
The indoor photos are collected from three physical scenes: 1) the underground passage (rainy, morning);
2) the laboratory corridor (sunny, afternoon); 3) the underground parking garage (cloudy, evening).
For each pair of $<$angle, distance$>$ (i.e., $0^\circ $, $15^\circ $, $30^\circ $), we take 100 photos for the testers wearing the 3D invisible cloaks to calculate the stealth attack success rate.
In particular, for the angle of $15^\circ $ and $30^\circ $, we capture 50 photos for the same angle on both sides ($ \pm 15^\circ $ and $ \pm 30^\circ $) of the camera.

\textbf{Conventional-cloak.} First, we evaluate the effectiveness of the Conventional-cloak for person stealth attacks.
The experimental results are presented in Appendix \ref{Appendix9.5}.

\textbf{The cloaks with single 3D physical transformation.} Then, we evaluate the impacts of different 3D physical constraints (wrinkle, radian, angle, and occlusion) on person stealth attack success rates.
For each type of invisible cloak (as shown in Figure \ref{fig:fig8}(b)-\ref{fig:fig8}(e)), we take 100 different photos for the testers wearing the cloak under the specific indoor and outdoor physical scenario of (3m, $15^\circ $).
Table \ref{tab:tab6} compares their stealth attack performances with that of the Conventional-cloak.
As shown in Table \ref{tab:tab6}, the invisible cloaks generated by considering 3D physical constraints can achieve higher stealth attack success rates than Conventional-cloak in both indoors and outdoors scenarios.
In outdoor conditions, the stealth attack success rates of these four 3D invisible cloaks reach 98\% (radian), 100\% (angle), 100\% (occlusion), and 98\% (wrinkle), respectively, which are significantly higher than that of the Conventional-cloak.
In indoor conditions, the stealth attack success rates of these four 3D invisible cloaks is 87\%, 94\%, 92\%, and 90\%, respectively, which are also significantly higher than that of the Conventional-cloak.

\begin{table}[htbp]
\renewcommand{\arraystretch}{0.7}
\linespread{0.8}
  \centering
  \footnotesize
  \caption{The attack results of different 3D invisible cloaks in the physical scenario of (3m, $15^\circ $)}
        \setlength{\tabcolsep}{0.2mm}{
    \renewcommand\arraystretch{-1}
    \begin{tabular}{m{9em}<{\centering}m{6em}<{\centering}m{6em}<{\centering}m{6.5em}<{\centering}}
    \toprule
    \multirow{4}[3]{*}{\textbf{\tabincell{c}{Invisible \\Cloaks types}}} & \multicolumn{3}{c}{\textbf{Attack success rate $R_{s}$}} \\
\cmidrule{2-4}          & \textbf{Outdoors} & \textbf{Indoors} & \textbf{Average} \\
    \midrule[0.09em]
    Conventional-cloak & 90\%  & 84\%  & 87\% \\
    \midrule[0.05em]
    Radian-cloak & 98\%  & 87\%  & 92.50\% \\
    \midrule[0.05em]
    Angle-cloak & 100\% & 94\%  & 97\% \\
    \midrule[0.05em]
    Occlusion-cloak & 100\% & 92\%  & 96\% \\
    \midrule[0.05em]
    Wrinkle-cloak & 98\%  & 90\%  & 94\% \\
    \midrule[0.05em]
    Combined-cloak & 100\%  & 95\%  & 97.5\% \\
    \bottomrule
    \end{tabular}%
    }
  \label{tab:tab6}%
\end{table}%

As shown in Table \ref{tab:tab6} (and Table \ref{tab:tab5} in Appendix \ref{Appendix9.5}), the stealth attack success rates outdoors are generally higher than that indoors.
The reason is that, the lighting conditions in outdoor photos are better than those photos collected indoors, which makes the YOLO v2 detector more easily and accurately capture the generated adversarial perturbations.
In addition, most of the Inria training images that used to train the adversarial patches are collected outdoors.
As a result, when the patches are printed on the real clothes, these generated 3D invisible cloaks can perform better under the outdoors physical scenarios.

Further, we evaluate the generated 3D invisible cloaks under more complex scenarios, such as the cloak is wrinkled and sheltered, the cloak faces the camera from different angles, and the cloak appears some stereoscopic radians.
In this paper, the test images are collected at various different angles of multiple people, and those testers are with different heights and body shapes.
Therefore, according to the above experiments (\textit{e.g.}, Table \ref{tab:tab6}), the proposed person stealth attack is resistent to the physical transformations of ``Radian'' and ``Angle''. Next, for Occlusion-cloak and Wrinkle-cloak, we cover random parts of the Occlusion-cloak, and make different degrees of wrinkles on Wrinkle-cloak, to further evaluate whether the cloak is resistent to the constraints of ``Occlusion'' and ``Wrinkle'', respectively.

\begin{figure}[h]
  \centering
  \footnotesize
  \includegraphics[width=2.8in]{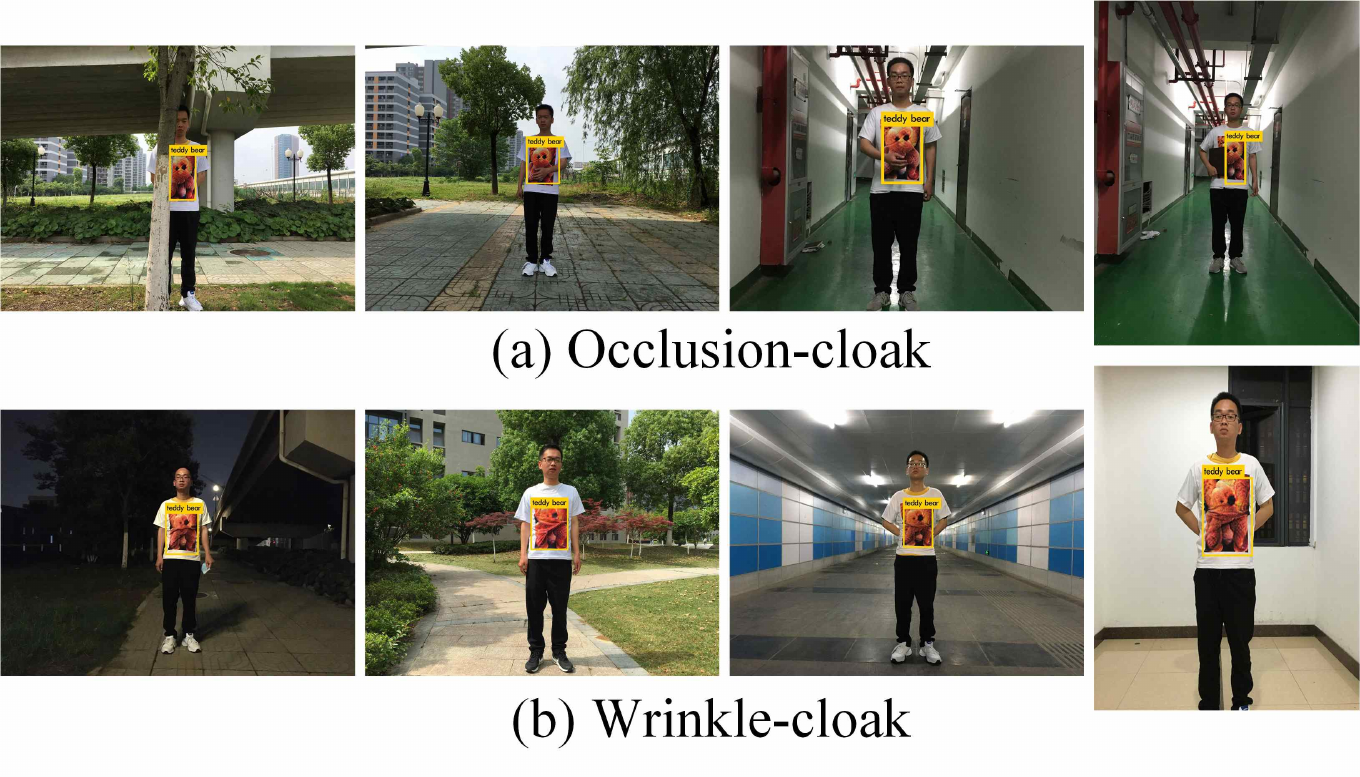}\\
  \caption{Examples of person stealth attacks with Occlusion-cloak and Wrinkle-cloak under more complex physical scenes.}
  \label{fig:fig15}
\end{figure}

Some attack examples of generated Occlusion-cloak and Wrinkle-cloak under these more complex physical conditions are shown in Figure \ref{fig:fig15}.
It is shown that, even part of the cloak has been occluded by a tree, a hand or a laptop (Figure \ref{fig:fig15}(a)), or wrinkled (Figure \ref{fig:fig15}(b)), the person wearing the cloak can still evade the detection of YOLO v2 in various indoor and outdoor physical scenarios.
The experimental results demonstrate that, in addition to the ``Radian'' and ``Angle'' constraints, the proposed person stealth attack is also robust to the constraints of ``Occlusion'' and ``Wrinkle'', in which the 3D invisible cloaks can achieve the stealth attack success rate of 81\% (Occlusion) and 80\% (Wrinkle), respectively.

\textbf{Combined-cloak.} Finally, we evaluate the effectiveness of the Combined-cloak (as shown in Figure \ref{fig:fig8}(f)), which can achieve the person stealth attack success rate of 86.56\% in digital domain.
The maximum and average stealth attack success rates of Combined-cloak at different distances are shown in Table \ref{tab:tab8}.
The average attack success rate (indoors \& outdoors) at each distance (2m, 3m, 4m) is the average of stealth attack success rates of Combined-cloak at three different angles ($0^\circ$, $15^\circ$, $30^\circ$).
Under those outdoor physical conditions, the maximum stealth attack success rate of Combined-cloak at different distances (2m, 3m, 4m) is 95\%, 100\% and 92\%, respectively.
The average attack success rate of Combined-cloak at 3m is high up to 93.33\%.
Under those indoor scenes, the maximum attack performance of Combined-cloak at different distances (2m, 3m, 4m) is 92\%, 99\% and 80\%, respectively. The average attack performance of Combined-cloak at 3m is 94.67\%.
As expected, the 3D cloak that considers all the four physical constraints performs better than the conventional cloak.
Besides, compared to those cloaks that only consider one constraint, the Combined-cloak is more effective in attacking the object detectors.
For example, as shown in Table \ref{tab:tab6}, in the attack scene of (3m, $15^\circ $), the average attack success rate of Combined-cloak is high up to 97.5\%, which is higher than those cloaks that only consider a single physical constraint.

\begin{table}[htbp]
\renewcommand{\arraystretch}{0.7}
\linespread{0.7}
  \centering
  \footnotesize
  \caption{Attack results of Combined-Cloak at different distances}
  \setlength{\tabcolsep}{0.5mm}{
    \renewcommand\arraystretch{0.5}
    \begin{tabular}{m{5em}<{\centering}m{5em}<{\centering}m{5em}<{\centering}m{5em}<{\centering}m{5em}<{\centering}}
    \toprule
    \multirow{2}[7]{*}{Distances} & \multicolumn{2}{c}{\textbf{\tabincell{c}{Attack success rate \\outdoors}}} & \multicolumn{2}{c}{\textbf{\tabincell{c}{Attack success rate \\indoors}}} \\
\cmidrule{2-5}          & Maximum & Average & Maximum & Average \\
    \midrule
    2m    & 95\%  & 93\% & 92\%  & 83\% \\
    \midrule
    3m    & 100\% & 93.33\% & 99\%  & 94.67\% \\
    \midrule
    4m    & 92\%  & 69.33\% & 80\%  & 61\% \\
    \bottomrule
    \end{tabular}%
    }
  \label{tab:tab8}%
\end{table}%

Note that, as shown in Table \ref{tab:tab8} (and Table \ref{tab:tab5} in Appendix \ref{Appendix9.5}), the attack performances of Combined-cloak (and Conventional-cloak) at 3m are better than that of 2m.
This is because, our adversarial patches are trained on the Inria training set, and most people in these Inria images are approximately at the same distance (3m) as the testers in our collected photos.
As a result, the generated patches can achieve better attack performances at such distance (3m).

\begin{figure}[htbp]
  \centering
  \includegraphics[width=3.35in]{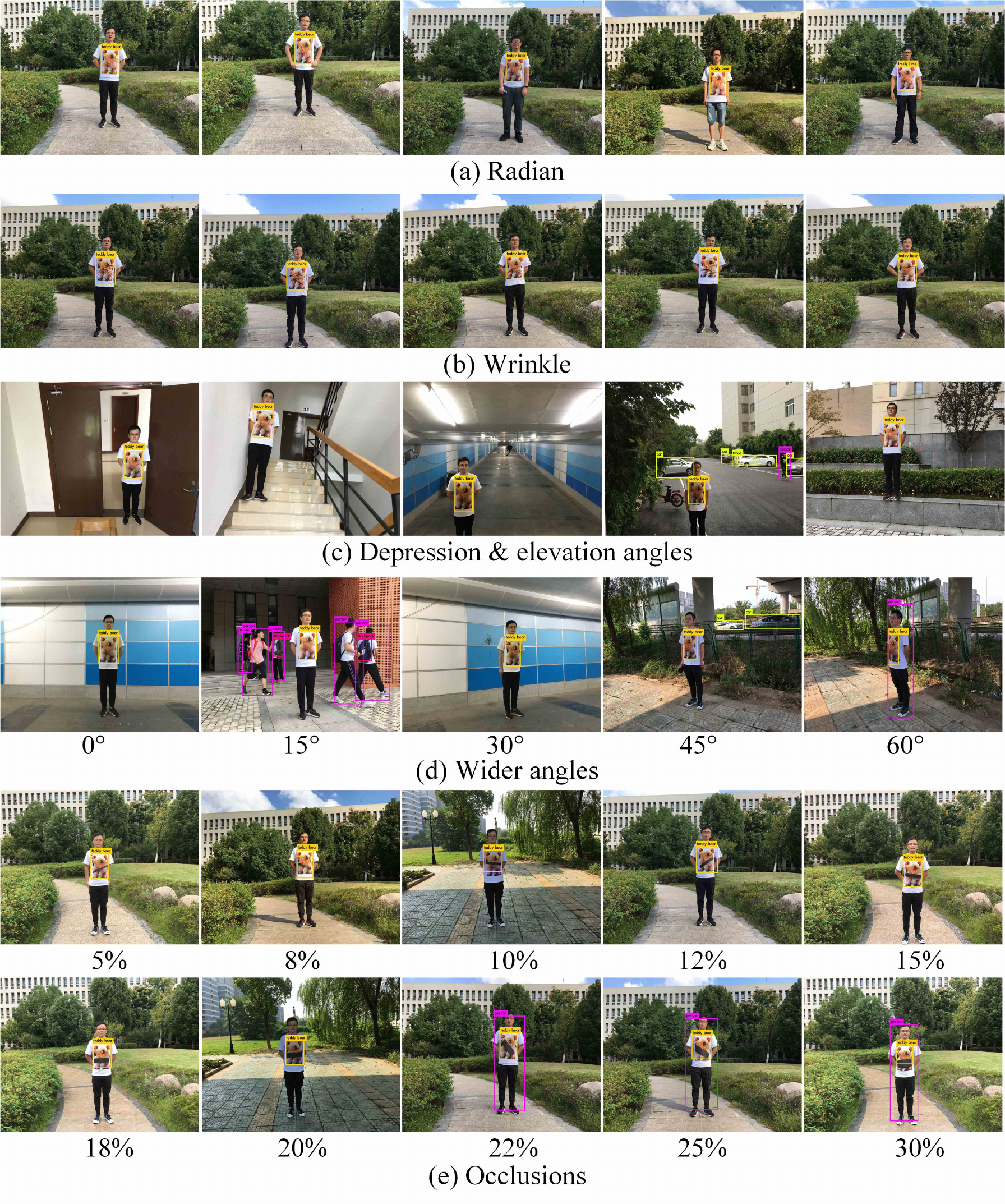}\\
  \caption{Attack results of Combined-cloak under more difficult/complex scenes.}
  \label{fig_com}
\end{figure}

Similarly, we evaluate the performance of Combined-cloak under more difficult scenarios, and some attack examples are presented in Figure \ref{fig_com}.

\textbf{-Radian.} First, we evaluate the robustness of Combined-cloak to the constraint ``Radian'', and some attack results are presented in Figure \ref{fig_com}(a).
It is shown that, different attackers wear the Combined-cloak can attack successfully.
This indicates that, the Combined-cloak is feasible for attackers that have different shapes of bodies (thin, medium or fat).

\textbf{-Wrinkle.} Then, for the physical constraint ``Wrinkle'', the generated Combined-cloak is also effective and robust.
Figure \ref{fig_com}(b) shows some attack examples, and the Combined-cloak performs well even there have different degrees of wrinkles on it. The attacker wearing such invisible cloak can successfully fool the object detector and achieve stealth.

\textbf{-Depression and elevation angles.} Next, in addition to these horizontal angles, the Combined-cloak is also robust under those depression and elevation angles.
Figure \ref{fig_com}(c) shows shows attack examples of Combined-cloak at depression and elevation angles under various physical conditions (indoors and outdoors). It is shown that, the Combined-cloak performs well when the camera (\textit{i.e.}, the object detector) captures the photos from top down or bottom up.

\textbf{-Wider angles.} Further, we explore the attack success rate of Combined-cloak under those wider angles.
Specifically, for each attack angle ($0^\circ$, $15^\circ$, $30^\circ$, $45^\circ$ and $60^\circ$), we collect 100 photos (testers wearing the cloak) at the distance of 3m under indoor and outdoor physical scenes.
Here, the angle is defined as $0^\circ$ when the camera is facing the testers. When the camera is deflected from the testers, the angle changes from $0^\circ$ to $60^\circ$.
Some attack examples are presented in Figure \ref{fig_com}(d).
The attack performance of Combined-cloak under different angles is 100\% ($0^\circ$), 95\% ($15^\circ$), 85\% ($30^\circ$), 46\% ($45^\circ$) and 0\% ($60^\circ$), respectively.
The attack success rate drops as the attack angle increases from $0^\circ$ to $60^\circ$.
This is because, the added perturbations on the cloak cannot be completely captured by the detector at a large angle (as shown in the last photo of Figure \ref{fig_com}(d)), which degrades the person stealth attack performance.

\textbf{-Different sizes of occlusion areas.} We make several black stickers of different sizes, which account for 5\%, 8\%, 10\%, 12\%, 15\%, 18\%, 20\%, 22\%, 25\%, 30\% of total area (29cm*43cm) of the printed patch.
In the experiments, these black stickers are pasted on a random area of the patch, and we take photos for the testers to evaluate the person stealth attack performance.
Some attack results are shown in Figure \ref{fig_com}(e), and the content below each picture represents the occlusion area on the cloak.
It is shown that, even the patch is occluded by 20\% of the area, the generated 3D cloak can still fool the object detector under various physical scenes.
The generated Combined-cloak fails when the occluded area increases to 22\% or larger.
The reason is that, when launching the stealth attacks against the object detector, all these adversarial perturbations on a cloak work together.
If a large area on the cloak is covered, these remaining perturbations (i.e., uncovered) will not be sufficient to cause the target detector to make incorrect predictions.
However, as shown in Figure \ref{fig:fig15}(a), for the Occlusion-cloak that only considers the constraint ``Occlusion'', it still works when the cloak has been sheltered by a larger object, such as a notebook or one hand.

Finally, in Appendix \ref{Appendix_9.6}, we demonstrate the generalization and convergence of the proposed person stealth attack.

 \vspace{-0.25cm}
\subsubsection{Dynamic Physical Stealth Attacks}
\label{sec:sec6.3.2}
\begin{figure}[!htbp]
  \centering
  \includegraphics[width=3.2in]{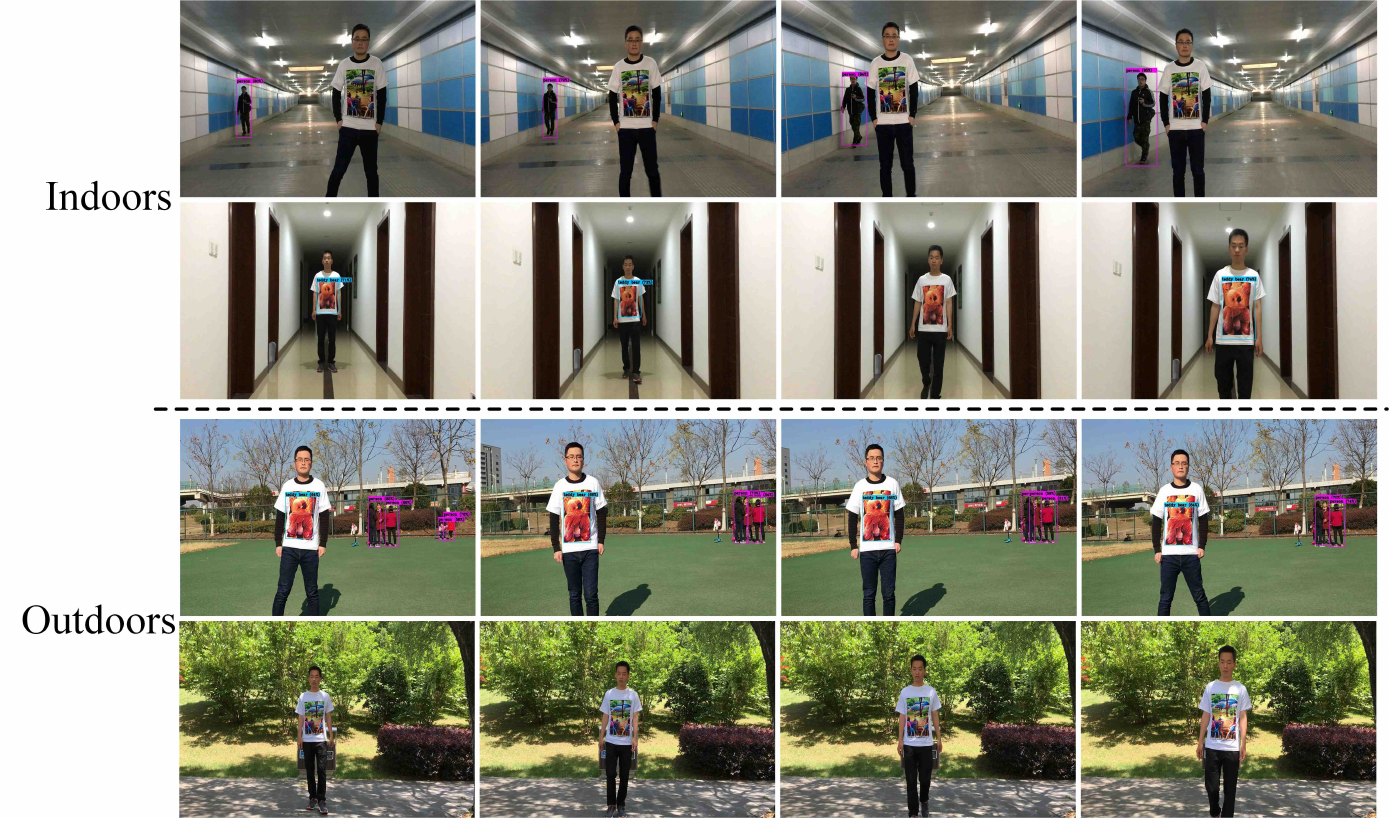}\\
  \caption{Some attack results of dynamic person stealth attacks.}
  \label{fig:fig11}
\end{figure}

For dynamic person stealth attacks, the continuous video frames of testers who wear the invisible cloaks are submitted to YOLO v2 detector.
The testers are moving, and the shapes of the invisible cloaks, the angles and distances they facing the camera are changing dynamically, which cause the person stealth attacks more difficult.
In our experiments, we record videos from the moving persons that wear the Conventional-cloak and the Disappeared-cloak (as discussed in Section \ref{sec:sec7}).
10 videos (5$\sim$30 seconds per video) are collected for each invisible cloak from various indoor and outdoor scenarios.
The experimental results show that, the Conventional-cloak can achieve the stealth attack success rate of 76.25\% indoors and 77\% outdoors, respectively.
The attack performance of Disappeared-cloak is 69.34\% indoors and 75.53\% outdoors, respectively.
This means that under those more realistic physical scenarios (dynamic videos), where the people are moving and their postures are changing dynamically, the 3D invisible cloaks generated by our proposed method are still robust for person stealth attacks.
Figure \ref{fig:fig11} shows some dynamic person stealth attack results. The person in these continuous video frames can successfully evade the detection of YOLO v2.

 \vspace{-0.25cm}
\subsection{Comparison with Existing Stealth Attacks}
\label{sec:sec6.4}
 \vspace{-0.25cm}

In this section, we compare the proposed person stealth attack with existing stealth attacks \cite{yang2018building,thys2019fooling}. Note that, there are several insurmountable difficulties when comparing the proposed 3D stealth attacks with the existing stealth attacks under the same condition. First, the two existing stealth attacks are separately aiming at Tiny YOLO (also called YOLO-LITE \cite{huang2018yolo}) \cite{yang2018building} and YOLO v2 \cite{thys2019fooling} detector, which makes the target detectors of these three stealth attacks different. Second, the existing stealth attack works \cite{yang2018building,thys2019fooling} did not discuss the size of the generated adversarial patch, as well as the distance and angle of stealth attacks, which cause their experiments lack detailed statistical experimental results in digital domain and under specific physical settings.
Therefore, we only select their highest stealth attack success rates in digital domain and real physical world for comparison.

Since the work \cite{yang2018building} has not provided the source codes of their implementations, we cannot obtain or generate the adversarial patch with their proposed attack method. Therefore, we directly use the attack results (digital and in physical world) that reported in their paper for comparison.
For the work \cite{thys2019fooling}, we use the adversarial patch in their provided project for comparison.
In digital domain, we paste this patch on the people in Inria test images for evaluation.
In the real world, we print it on real clothes to generate the invisible cloak (as shown in Figure \ref{fig:fig8}(l)), to evaluate its performance under physical conditions.
For fair comparisons, under the same experimental settings (indoor and outdoor scenes) as in this paper, we take the same number (100) of photos for the testers wearing such cloak at each pair of <distance, angle>. The comparison of stealth attack results are shown in Table \ref{tab:tab7}.

\begin{table}[htbp]
\linespread{0.7}
  \centering
  \footnotesize
  \caption{Comparison with existing person stealth attacks}
         \setlength{\tabcolsep}{0mm}{
    \renewcommand\arraystretch{-1}
    \begin{tabular}{m{7.5em}<{\centering}m{5em}<{\centering}m{5em}<{\centering}m{7.5em}<{\centering}}
    \toprule
   \textbf{Attack settings} & \textbf{\cite{yang2018building}} & \textbf{\cite{thys2019fooling}} & \textbf{Our method} \\
    \midrule[0.09em]
    Target objectors & Tiny YOLO & YOLO v2& YOLO v2\\
    \midrule[0.05em]
    Digital attack success rate & 100\% & 47.31\% & 86.56\% \\
    \midrule[0.05em]
    Physical attack success rate & 72\%  & 17\%  & 100\% \\
    \bottomrule
    \end{tabular}%
    }
  \label{tab:tab7}%
\end{table}%
 \vspace{-0.25cm}

\textbf{Physical attack success rate.} It is shown that, since we generate the invisible cloaks with those optimal adversarial patches and consider the 3D wearable transformations, the attack performances of our invisible cloaks are much better than the existing works \cite{thys2019fooling, yang2018building} in real physical world.
The physical stealth attack success rate of 3D invisible cloak in this paper is up to 100\%, while the attack success rate of work \cite{yang2018building} and \cite{thys2019fooling} in physical world is only 72\% and 17\%, respectively.

\textbf{Digital attack success rate.} In digital domain, the stealth attack success rate of this paper (86.56\%) is slighter lower than that of work \cite{yang2018building} (100\%), but much higher than the attack success rate (47.31\%) of work \cite{thys2019fooling}. This is because, the work \cite{yang2018building} only took 50 photos for the same single person standing in the middle of the camera at several scenes, as their test images in digital domain.
However, to comprehensively evaluate the performance of an adversarial patch, the patch should be tested on a larger number of images that contains various indoor and outdoor scenes.
In this paper, we evaluate the digital attack performances of our generated patches on the Inria dataset \cite{dalal2005histograms}.
The 288 images in Inria dataset includes more people (602 in total), and these people appear at different distances and angles from the camera in various scenes.
Moreover, some Inria images \cite{dalal2005histograms} have only captured a part of a person's body, which greatly increase the difficulty of stealth attacks.
Therefore, the attack performance of this paper is slightly lower than that of work \cite{yang2018building} in digital domain.

Note that, the success rate of stealth attack in real physical world (100\%) is higher than that in digital domain (86.56\%).
The reasons are as follows.
First, in digital domain, the adversarial patch is scaled and pasted on the person in Inria images to implement the attack.
However, the people in Inria images are quite small, which causes the patches pasted on these images are so tiny that the patches cannot be recognized by YOLO v2, thus resulting in a relatively lower attack success rate in digital domain.
Second, these 3D invisible cloaks are generated by considering 3D physical constraints, which do not exactly match the digital simulation scenarios.
Finally, compared to these images in digital domain (Inria dataset \cite{dalal2005histograms}), the testers who wear the 3D invisible cloak in the collected physical photos are complete, and there are relatively fewer interferences in the background.

\vspace{-0.25cm}
\section{Disappeared 3D Stealth Attacks}
\label{sec:sec7}
 \vspace{-0.25cm}
In this section, we aim to generate such an invisible cloak that can make a person completely ``disappeared''.
As discussed in Section \ref{sec:sec6}, the six different types of invisible cloaks (as shown in Figure \ref{fig:fig8}(a)-\ref{fig:fig8}(f)) can make a person evade the detection of YOLO v2, and the patterns on the cloaks will be detected as ``teddy bear''.
However, in some attack scenarios where there has high security levels, such as banks and customs, the ``teddy bear'' on the invisible cloak may arouse unnecessary attentions, since the teddy bear is not supposed to be appeared in these scenes.
Therefore, the ideal stealth attack should be that, the generated 3D cloak will not be detected as any object, i.e., the person is completely ``disappeared'' under the detection of YOLO v2 detector.
In this way, an attacker can covertly achieve his malicious purpose without being noticed at all.

\begin{figure}[htbp]
  \centering
  \includegraphics[width=2.3in]{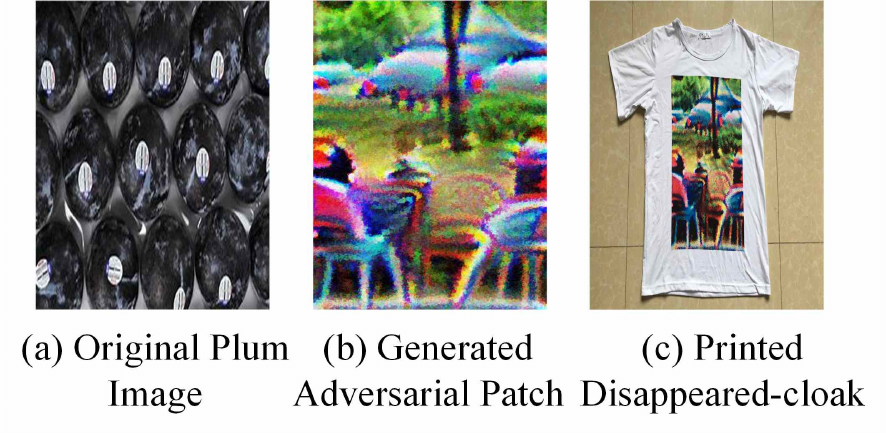}\\
  \caption{The original plum image, the generated Disappeared-patch, and the printed Disappeared-cloak.}\label{fig:fig12}
\end{figure}

In this paper, we not only reduce the confidence of the class ``person'' in the bounding boxes of an input image, but also reduce the probabilities that other classes being detected in the adversarial patch.
In this way, the generated adversarial patch will not be detected as any object.
We use a plum image as an example and modify its objective loss function. The original plum image and its generated adversarial patch, printed invisible cloak under the conventional transformations are shown in Figure \ref{fig:fig12}.
The generated adversarial patch no longer looks like the ``teddy bear''.
For the convenience of description, the invisible cloak printed with the adversarial patch shown in Figure \ref{fig:fig12}(b) will be referred to as the Disappeared-cloak, as shown in Figure \ref{fig:fig12}(c).
Experimental results in Section \ref{sec:sec6.3} show that, the invisible cloaks generated by considering the 3D transformations $U$ have much better stealth attack performance than that generated by only considering the conventional transformations $R$.
Therefore, in this section, we generate the Disappeared-cloak by considering the conventional transformations $R$ only (which has a lower performance).
If the invisible cloak generated by considering conventional transformation $R$ can successfully make the person disappear, the cloaks generated by considering the 3D transformations $U$ can also make the person disappear.

\begin{figure}[htbp]
  \centering
  \includegraphics[width=2.7in]{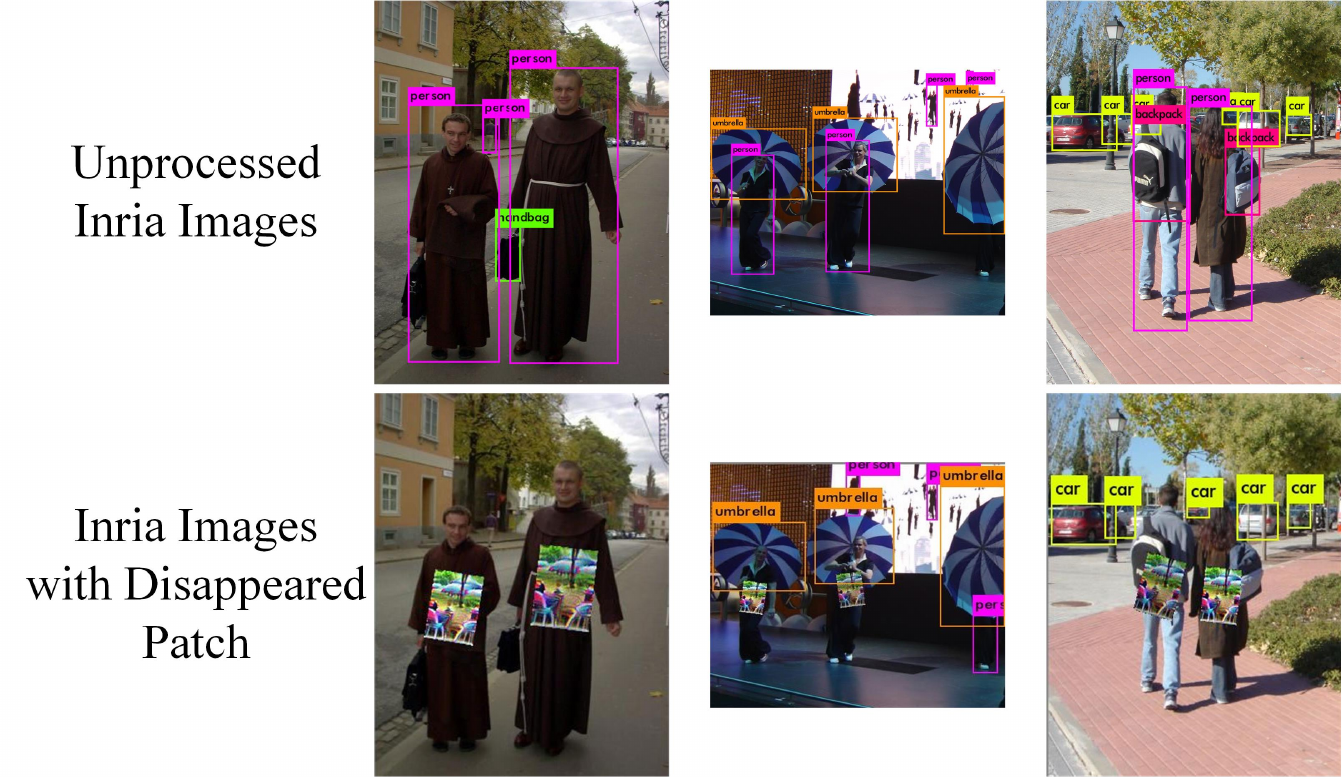}\\
  \caption{Some attack results of Disappeared-patch on Inria test set.}\label{fig:fig13}
\end{figure}

Figure \ref{fig:fig13} shows some experimental results of the stealth attack on the Inria test set with this Disappeared-patch. The first row are the detection results of the unprocessed Inria images, and the second row shows the detection results after pasting the Disappeared-patch on the person. It is shown that the person pasted with such patch can evade the detection of YOLO v2, and will not be recognized as any objects. However, the stealth attack success rate of this Disappeared-patch on the Inria test set is only 71.92\%, which is lower than that of Conventional-patch with the ``teddy bear'' pattern. This is because the adversarial patch with the shape of ``teddy bear'' can achieve the optimal stealth attack performance against the YOLO v2 detector.
Although the Disappeared-patch can make the people not be recognized as any objects, the patch has been far away from the optimal performance of the stealth attacks, which results in the decline in its stealth attack performance.

\begin{figure}[htbp]
  \centering
  \setlength{\belowcaptionskip}{-0.25cm}
  \includegraphics[width=2.5in]{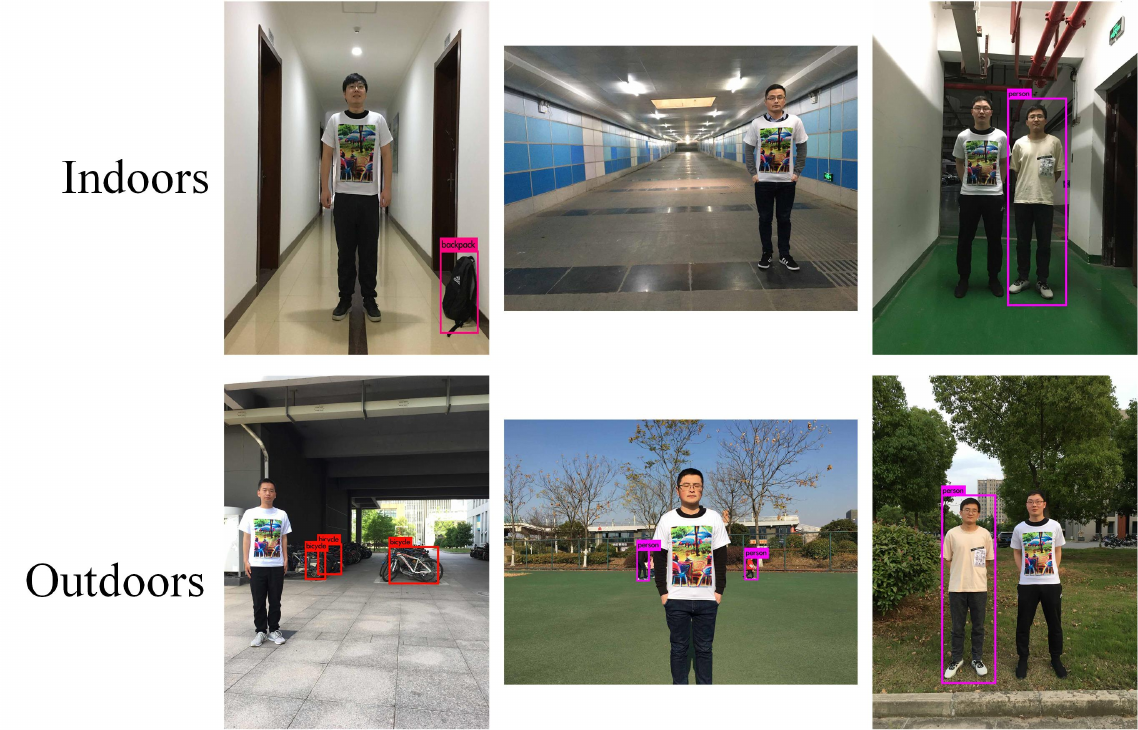}\\
  \caption{Some attack results of Disappeared-cloak in real world.}\label{fig:fig14}
\end{figure}

In addition, we print the Disappeared-patch on real clothes to generate the 3D wearable invisible cloak, and evaluate its attack performance in real world.
Figure \ref{fig:fig14} shows some stealth attack results of the Disappeared-cloak in real physical world. It is shown that the person wearing such Disappeared-cloak can evade the detection of YOLO v2 under indoors and outdoors conditions, and will not be detected as any objects.
For each pair of $<angle, distance>$, we take 100 photos for the testers wearing the Disappeared-cloak. The attack results show that, under the outdoor physical scenes, the average stealth attack success rate of Disappeared-cloak at the distance of 2m, 3m is 82.67\% and 89.00\%, respectively.
Under the indoor conditions, the Disappeared-cloak can achieve the average attack success rate of 62.33\% (2m) and 82.33\% (3m), respectively.
Similar to those ``teddy bear'' cloaks, the performances of Disappeared-cloak are relatively lower when the attack distance increases to 4m. The reasons are the same as discussed in Section \ref{sec:sec6.3}.
The attack performances of Disappeared-cloak are lower than that of the Combined-cloak.
We believe that there is a restrictive relationship between the shape of the adversarial patch and its attack success rate. The adversarial patches with the shape of teddy bears can achieve the highest stealth attack performances, but they will be detected as ``teddy bear''. On the other hand, those patches that are not recognized as any objects have relatively lower stealth success rates.

 \vspace{-0.25cm}
\section{Conclusion}
\label{sec:sec8}
In this paper, we propose a novel physical person stealth attack, where anyone wearing the generated 3D invisible cloaks can evade the detection of object detectors (YOLO v2) and become ``invisible''.
Compare to these existing stealth attacks, for the first time, this paper considers and addresses the impacts of those challenging 3D physical constraints (radian, wrinkle, angle, and occlusion) on person stealth attacks.
We print the adversarial patches on real clothes to launch the person stealth attacks targeting complex 3D physical space instead of 2D plane or digital simulations.
Moreover, through the methodical experimental designs, this paper explores which type of input image (e.g., color, shape of contained objects) can generate the optimal 3D invisible cloak.
A systematic evaluation framework is also proposed. Experimental results show that, the generated 3D invisible cloaks are robust and feasible even under those complex physical scenarios.
The proposed person stealth attack can achieve the attack success rate of 86.56\% in digital domain, while the attack performance in physical world is up to 100\% (static) and 77\% (dynamic), respectively.
This work demonstrates the vulnerability of advanced person detectors to those realistic physical attacks, and highlights the necessity of developing more robust object detectors.
Our future works will focus on the reasons behind such vulnerabilities and the countermeasures.
\vspace{-0.2cm}

\bibliographystyle{unsrt}%
\bibliography{ref}

\section{Appendix} \label{Appendix}

\subsection{Experimental Setup} \label{Appendix_F1}

\subsubsection{Dataset}
In this paper, the generation of adversarial patch and the evaluation of person stealth attack in the digital domain are conducted on the Inria dataset \cite{dalal2005histograms}.

\textbf{Inria Dataset \cite{dalal2005histograms}.} The Inria dataset is a set of labeled images of standing and walking people, which collected by Navneet Dalal and Bill Triggs in their study of pedestrian detection \cite{dalal2005histograms}.
The images of the dataset are collected from GRAZ-01 dataset \cite{opelt2004graz}, personal photos, and Google website, \textit{etc}.
The poses of people and the lighting conditions in Inria dataset are comprehensive, and people in those Inria images appear in various background, which are very close to the real physical scenarios.
The Inria training set consists of 614 positive samples containing people, and 1,218 negative samples without any people \cite{dalal2005histograms}. The Inria test set consists of 288 positive samples that contain persons and 453 negative samples without any person \cite{dalal2005histograms}.
In this paper, the adversarial patch is trained on the 614 positive images of Inria training set, and the evaluation of attack performance in the digital domain is conducted on the 288 positive images of the Inria test set.

In physical world, we use mobile phones (iPhone 6S) to take photos for the testers wearing the 3D invisible cloaks in various physical scenarios. Then, we submit these collected images to YOLO v2 \cite{redmon2017yolo9000} for person detection. For each specific physical setting $<$angle, distance$>$, we take 100 different photos under the indoor and outdoor scenarios.
In this paper, the number of tested images is more than 5,000, with a total size of 15GB, which will be made available after the blind review process.

\subsubsection{YOLO v2 Object Detector} \label{Appendix_F2}
The proposed person stealth attack targets at YOLO v2 detector \cite{redmon2017yolo9000}. The pre-trained weight file \textit{yolov2.weight} is obtained from the official website of YOLO v2\footnote{https://pjreddie.com/darknet/yolov2/} for the evaluation of the proposed attack.

\textbf{YOLO v2 \cite{redmon2017yolo9000}}. The YOLO v2 detector is an one-stage object detector, which can mark the positions of multiple objects in an input image and predict their probability distributions among 80 classes within one step \cite{redmon2017yolo9000}.
For each input image, first, YOLO v2 divides it into $S*S$ square grid cells of the same size.
Second, the detector generates $B*B$ different bounding boxes for each grid cell. The predictions of each bounding box includes \cite{redmon2017yolo9000}: 1) the position $(x,y)$, the width and height $(w,h)$ of this bounding box; 2) the probability $P_{obj}$ that this bounding box contains an object; 3) the probability distribution of $m$ class labels for the object in this bounding box, $P(class) = [P(class1), P(class2),..., P(classm)]$ \cite{redmon2017yolo9000}.
Third, YOLO v2 calculates the product of the probability $P_{obj}$ and the maximum class probability in $P(class)$. If the product is less than the detection threshold, the corresponding bounding box will be discarded. Otherwise it will be retained \cite{redmon2017yolo9000}.
Finally, for each object, the YOLO v2 uses the non-maximum suppression (NMS) \cite{neubeck2006efficient} algorithm to remove the redundant bounding boxes and retain the one with the highest confidence.

\subsubsection{Evaluation Metric}
A successful person stealth attack is that the person can evade YOLO v2 detector in digital domain and in real physical world. Therefore, we define a metric, named \textit{the attack success rate} $R_{s}$, to evaluate the performance of the proposed person stealth attack:
\begin{equation}
  {R_s} = {{{N_{undect}}} \over {{N_{all}}}} \times 100\% {\kern 1pt}
\end{equation}
where ${N_{all}}$ is the total number of person among all test images, and ${N_{undect}}$ is the number of undetected person.

\subsection{Patches Generated from Images of Different Colors} \label{Appendix_F3}
In this part, to explore whether the color of an original image and the presence of objects in an image affect the attack performance of the generated adversarial patches, we evaluate the patches that generated from original images of different colors (pure-color and mixed-color).

\begin{figure}[!htbp]
  \centering
  \includegraphics[width=3in]{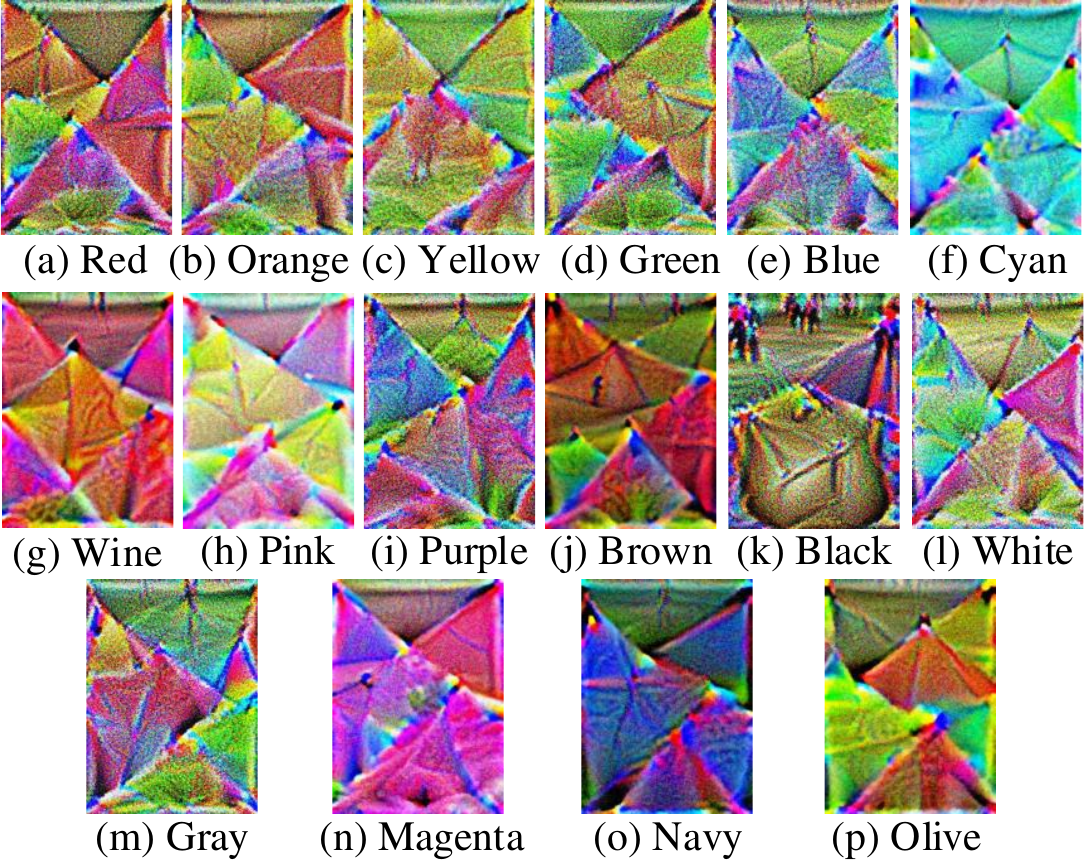}\\
  \caption{Adversarial patches generated from Pure-color images.}\label{fig:fig3}
\end{figure}

\textbf{Pure-color images.} Figure \ref{fig:fig3} shows the adversarial patches generated from the pure-color images. It is shown that, these generated adversarial patches all look like ``kite''.
The reason is that, the adversarial patches generated from different original images are all optimized by the same objective function (Equation (\ref{eq:eq5})).
In addition, for each pure-color image, the pixel value of each point is completely the same, which causes these pixel values on a pure-color image are uniformly distributed.
As a result, when these pure-color images are optimized by the same function, their generated patches will converge to a similar shape, except for the different brightness in some regions.
When the patches are pasted on the person, the person will be incorrectly identified as the ``kite'' by YOLO v2 and successfully achieve stealth.

Table \ref{tab:tab1} reports the attack results of these adversarial patches that presented in Figure \ref{fig:fig3}.
It is shown that:
1) the stealth attack performances of these patches generated by pure-color images are relatively poor;
2) these adversarial patches visually look similar, and will be detected as ``kite'' by YOLO v2;
3) among all these pure colors, the adversarial patch generated from the pure black image achieves the best stealth attack performance, with the attack success rate of 63.98\%.
This is because, unlike the adversarial patches generated from other pure colors, there is a dark area at the bottom of this patch. Such area will be recognized by YOLO v2 as ``backpack'', which causes this patch achieves a relatively high attack success rate;
4) except for the pure black, the attack success rates of adversarial patches generated by the other kinds of pure-color images are around 50\%. The highest, average, lowest stealth attack success rate among the other pure colors are 52.16\%, 50.46\%, and 49.30\%, respectively.

In our experiments, we generate one adversarial patch per color.
Under the same experimental settings, for each pure color, repeat the proposed adversarial patches generation method for $N$ times can output $N$ different adversarial patches.
However, the overall shapes of these generated patches are similar (``kite''), and their attack performances are at the same level.
For example, we generate 10 different adversarial patches for the pure-color red, and the average attack success rate of its generated 10 adversarial patches on Inria test set \cite{dalal2005histograms} is 51.72\%.
This indicates that the adversarial patches generated from the pure-color images are not optimal.
Therefore, this paper only generates one adversarial patch for each pure color.

\begin{table}[htbp]
  \centering
  \footnotesize
  \caption{The attack results of adversarial patches generated from Pure-Color images in digital domain}
    \begin{tabular}{cllcll}
    \toprule
    \textbf{\tabincell{c}{Color of \\original \\image}} & \multicolumn{1}{c}{\textbf{\tabincell{c}{Patch \\shape}}} & \multicolumn{1}{c}{\textbf{\tabincell{c}{Attack \\success \\rate $R_{s}$}}} & \textbf{\tabincell{c}{Color of \\original \\image}} & \multicolumn{1}{c}{\textbf{\tabincell{c}{Patch \\shape}}} & \multicolumn{1}{c}{\textbf{\tabincell{c}{Attack \\success \\rate $R_{s}$}}} \\
    \midrule
    \textbf{Red} & \multicolumn{1}{c}{Kite} & \multicolumn{1}{c}{51.65\%} & \textbf{Purple} & \multicolumn{1}{c}{Kite} & \multicolumn{1}{c}{50.59\%} \\
    \midrule
    \textbf{Orange} & \multicolumn{1}{c}{Kite} & \multicolumn{1}{c}{49.69\%} & \textbf{Brown} & \multicolumn{1}{c}{Kite} & \multicolumn{1}{c}{50.60\%} \\
    \midrule
    \textbf{Yellow} & \multicolumn{1}{c}{Kite} & \multicolumn{1}{c}{49.41\%} & \textbf{Black} & \multicolumn{1}{c}{Kite} & \multicolumn{1}{c}{63.98\%} \\
    \midrule
    \textbf{Green} & \multicolumn{1}{c}{Kite} & \multicolumn{1}{c}{49.33\%} & \textbf{White} & \multicolumn{1}{c}{Kite} & \multicolumn{1}{c}{51.98\%} \\
    \midrule
    \textbf{Blue} & \multicolumn{1}{c}{Kite} & \multicolumn{1}{c}{49.46\%} & \textbf{Gray} & \multicolumn{1}{c}{Kite} & \multicolumn{1}{c}{52.16\%} \\
    \midrule
    \textbf{Cyan} & \multicolumn{1}{c}{Kite} & \multicolumn{1}{c}{49.30\%} & \textbf{Magenta} & \multicolumn{1}{c}{Kite} & \multicolumn{1}{c}{50.49\%} \\
    \midrule
    \textbf{Wine} & \multicolumn{1}{c}{Kite} & \multicolumn{1}{c}{51.93\%} & \textbf{Navy} & \multicolumn{1}{c}{Kite} & \multicolumn{1}{c}{49.51\%} \\
    \midrule
    \textbf{Pink} & \multicolumn{1}{c}{Kite} & \multicolumn{1}{c}{50.78\%} & \textbf{Olive} & \multicolumn{1}{c}{Kite} & \multicolumn{1}{c}{50.16\%} \\
    \bottomrule
    \end{tabular}%
  \label{tab:tab1}%
\end{table}%

\textbf{Mixed-color images.} Then, we explore the stealth attack performances of those adversarial patches generated from mixed-color images.
For each mixed-color $c \in Color$ (as discussed in Section \ref{sec:sec6.2.2}), we select six mixed-color images that dominated by color $c$ to generate the adversarial patches, \textit{i.e.}, we generate 6 different adversarial patches for each mixed color.
In this way, a total of 60 different adversarial patches are generated from these mixed-color images.

\begin{figure}[htbp]
  \centering
  \footnotesize
  \includegraphics[width=2.4in]{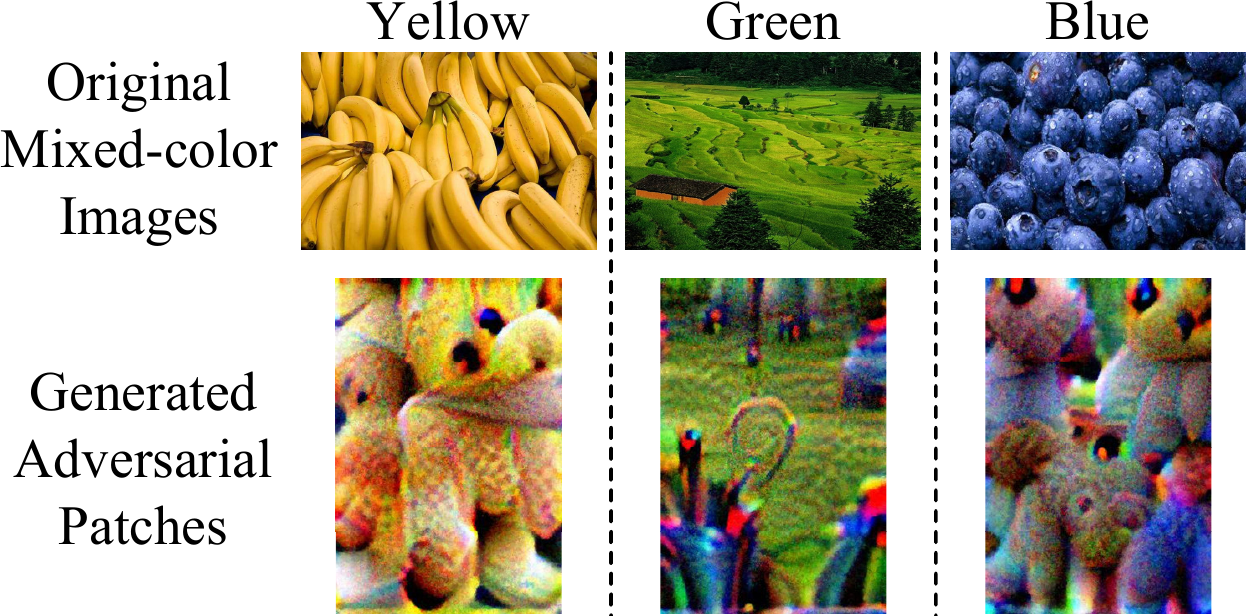}\\
  \caption{Some mixed-color images and their generated adversarial patches.}\label{fig:fig4}
\end{figure}

Table \ref{tab:tab2} shows the shapes of adversarial patches that generated by these 10 mixed colors, and their stealth attack success rates on Inria test set.
For each mixed color, the ``Patch shape'' represents the shape of the patch that has the best attack performance among all its 6 generated patches, and the ``Attack success rate $R_{s}$'' is the average stealth attack success rate of its 6 generated patches.
It is shown that, the stealth attack performances of those patches generated by different mixed-color images (contain some objects) are significantly different, and are generally higher than those patches that generated from pure-color images (contain no objects). The attack performances can be roughly divided into the following three levels. The mixed-color images that dominated by red, orange, yellow and brown achieve the best stealth attack performance on Inria test set, with average attack success rate of 82.69\%, 84.34\%, 83.90\%, and 82.83\%, respectively. The stealth attack performances of mixed-color images that dominated by purple and black are in the middle level, where the average stealth attack success rates are 79.47\% (purple) and 72.51\% (black), respectively. The mixed-color images dominated by green, blue, white, and gray have relatively lower stealth attack success rates, with the average attack success rates of 66.37\%, 42.96\%, 51.53\%, and 50.00\%, respectively.

Further, we find that the shape of a generated adversarial patch has a great impact on its stealth attack success rate. The shapes of adversarial patches in Table \ref{tab:tab2} can be divided into two classes: 1) Bears; 2) Non-Bears (``Kite'' and ``Others''). The adversarial patches with the shape of ``Bears'' have better attack performances, while the stealth attack success rates of the patches with the shape of ``Kite'' and ``Others'' are relatively low. Moreover, even for the same shape (``Bear''), the attack performances of these adversarial patches that generated from different colors of original images are quite different. This further demonstrates that the color of the original image has a great impact on the generated adversarial patch.

\begin{table}[!htbp]
  \centering
  \footnotesize
  \caption{The attack results of adversarial patches generated from mixed-color images in digital domain}
    \begin{tabular}{cccccc}
    \toprule
    \textbf{\tabincell{c}{Main color \\of mixed-\\images}} & \textbf{\tabincell{c}{Patch \\shape}} & \textbf{\tabincell{c}{Attack \\success \\rate $R_{s}$}} & \textbf{\tabincell{c}{Main color \\of mixed-\\images}} & \textbf{\tabincell{c}{Patch \\shape}} & \textbf{\tabincell{c}{Attack \\success \\rate $R_{s}$}} \\
    \midrule
    \textbf{Red} & Bear & 82.69\% & \textbf{Purple} & Bear & 79.47\% \\
    \midrule
    \textbf{Orange} & Bear & 84.34\% & \textbf{Brown} & Bear & 82.83\% \\
    \midrule
    \textbf{Yellow} & Bear & 83.90\% & \textbf{Black} & Others & 72.51\% \\
    \midrule
    \textbf{Green} & Bear & 66.37\% & \textbf{White} & Kite  & 51.53\% \\
    \midrule
    \textbf{Blue} & Kite  & 42.96\% & \textbf{Gray} & Others & 50.00\% \\
    \bottomrule
    \end{tabular}%
  \label{tab:tab2}%
\end{table}%

\subsection{The Printed 3D Invisible Cloaks} \label{Appendix9.4}

\begin{figure}[!htbp]
  \centering
  \includegraphics[width=3in]{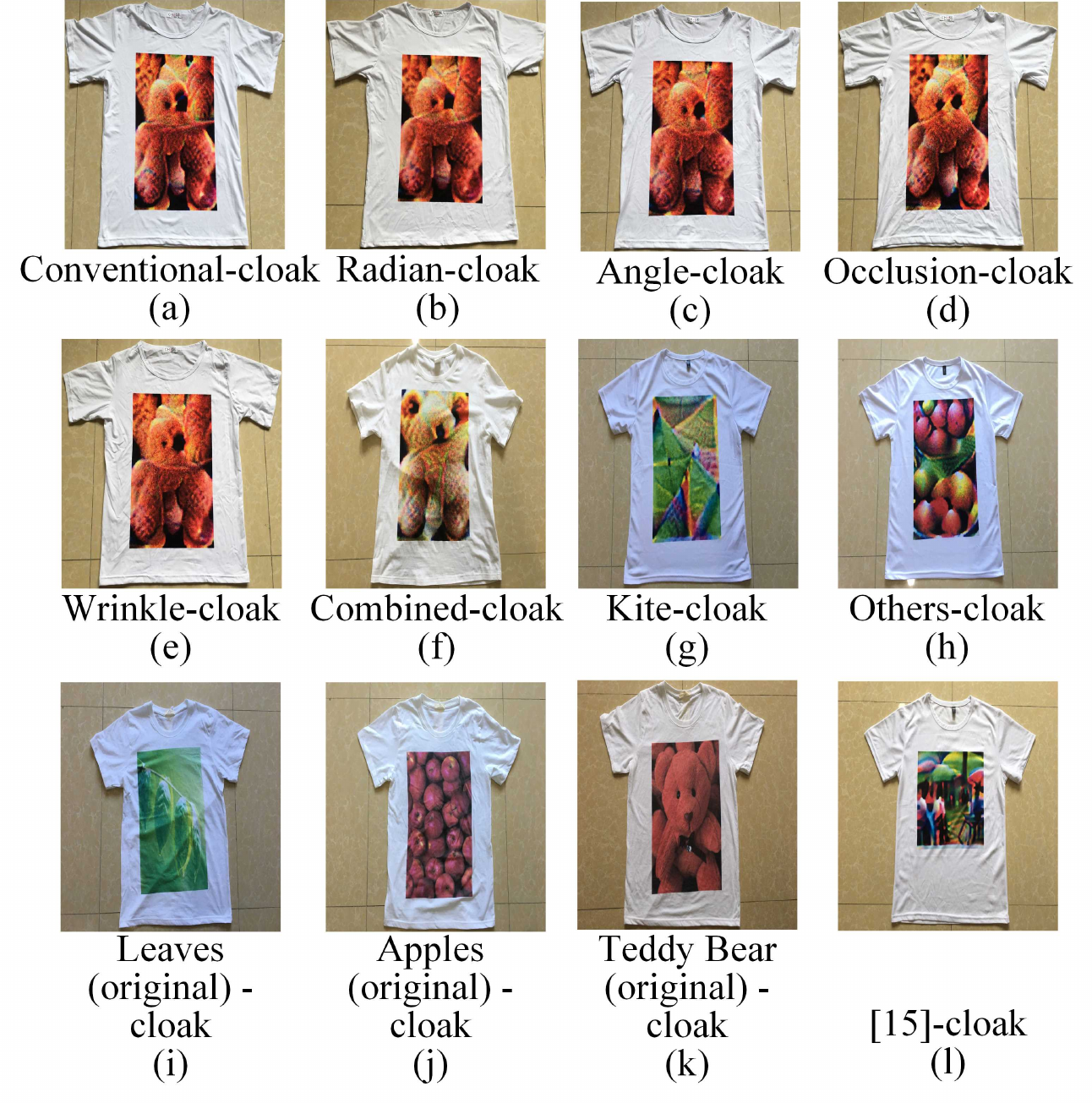}\\
  \caption{The printed 3D invisible cloaks in the experiments.}\label{fig:fig8}
\end{figure}

Figure \ref{fig:fig8} presents the printed 3D invisible cloaks in our experiments.
It is shown that, these 3D invisible cloaks look natural as ordinary T-shirts, thus will not arouse humans' suspicions.

\subsection{Attack Results of Conventional-cloaks in Real World} \label{Appendix9.5}

\begin{figure}[htbp]
  \centering
  \includegraphics[width=3.2in]{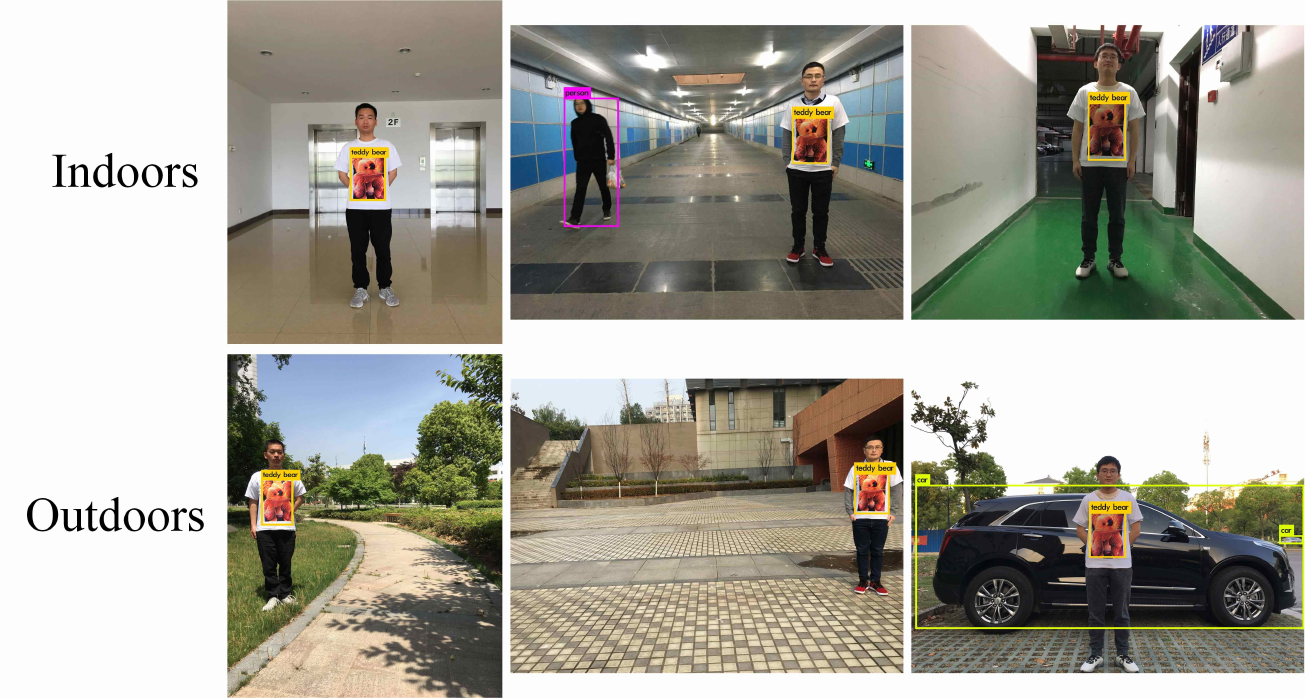}\\
  \caption{Examples of person stealth attacks in real physical world.}
  \label{fig:fig9}
\end{figure}

Some examples of person stealth attack in real physical world are presented in Figure \ref{fig:fig9}.
It is shown that the person wearing the Conventional-cloak can successfully evade the detection of YOLO v2 in indoors and outdoors scenarios.

In indoors and outdoors scenarios, we implement the person stealth attacks at different angles ($0^\circ $, $15^\circ $, $30^\circ $) and distances (2m, 3m, 4m).
The stealth attack results are shown in Table \ref{tab:tab5}.
It is shown that the generated cloak performs well at various angles in the distance of 2m and 3m, where the highest attack success rate of 2m and 3m reaches 100\% (indoors, $0^\circ $) and 99\% (outdoors, $0^\circ $), respectively. The performance of this invisible cloak at 4m with angle of $30^\circ $ is better than the same distance with smaller angles ($0^\circ $ and $15^\circ $).
This is because, the test photos collected at a large angle ($30^\circ$) are limited by the viewpoint of the camera, which causes the actual distance of the person in a collected image at (4m, $30^\circ $) is less than 4m.
Therefore, at the distance of 4m, the attack success rate at $30^\circ $ is higher than that at $0^\circ $ and $15^\circ $.

\begin{table}[htbp]
\renewcommand{\arraystretch}{0.7}
\linespread{0.8}
  \centering
  \footnotesize
  \caption{The attack results of Conventional-Cloak in real world}
  \setlength{\tabcolsep}{0mm}{
    \renewcommand\arraystretch{-1}
    \begin{tabular}{m{4.5em}<{\centering}m{3.5em}<{\centering}m{8.0em}<{\centering}m{8.0em}<{\centering}}
    \toprule
    \textbf{Distances} & \textbf{Angles} & \textbf{Attack success rate $R_{s}$ outdoors} & \textbf{Attack success rate $R_{s}$ indoors} \\
    \midrule[0.09em]
    \multicolumn{1}{c}{\multirow{3}[11]{*}{2 m}} & $0^\circ $   & 100\% & 99\% \\
\cmidrule{2-4}          & $15^\circ $   & 94\%  & 77\% \\
\cmidrule{2-4}          &$30^\circ $   & 62\%  & 45\% \\
    \midrule[0.05em]
    \multicolumn{1}{c}{\multirow{3}[11]{*}{3 m}} & $0^\circ $    & 99\%  & 97\% \\
\cmidrule{2-4}          & $15^\circ $   & 90\%  & 84\% \\
\cmidrule{2-4}          & $30^\circ $   & 98\%  & 91\% \\
    \midrule[0.05em]
    \multicolumn{1}{c}{\multirow{3}[11]{*}{4 m}} & $0^\circ $    & 0\%   & 1\% \\
\cmidrule{2-4}          &  $15^\circ $ & 5\%   & 15\% \\
\cmidrule{2-4}          & $30^\circ $   & 74\%  & 51\% \\
    \bottomrule
    \end{tabular}%
    }
  \label{tab:tab5}%
\end{table}%

In general, to solve the impact of the changes in distances on the attack performances of invisible cloaks, the adversarial patches are randomly scaled to simulate the changes of distance between the attackers and the object detectors \cite{thys2019fooling}.
In this paper, when generating an adversarial patch, we also perform the physical transformation ``scale'' on it during the conventional transformations $R$.
However, the stealth attack performance of Conventional-cloak degrades as the distance between the attackers and the camera increases (4m).
The reasons are as follows.
First, the patches printed on the clothes are too small ($29cm * 43cm$).
After reaching a certain distance, the adversarial perturbations added in the cloaks are difficult to be captured by the camera.
Second, the performance of the camera (iphone 6s) used in this paper is relatively poor, which cannot capture high resolution photos from long distances.
This causes those adversarial perturbations in an invisible cloak be ignored by the detectors at a long distance.
To address this problem, taking photos with a high performance camera or scaling the adversarial patch in proportion to the attacker's distance from the camera (rather than randomly scaled in this paper) can greatly improve the attack performances of the generated 3D invisible cloaks at large distances.

\subsection{Generalization \& Convergence of the Proposed Attack} \label{Appendix_9.6}
\textbf{Generalization of the proposed person stealth attack.} First, we demonstrate that the proposed person stealth attack method is generalizable. In other words, other types of generated adversarial patches are also effective, rather than only those ``teddy bear'' patches.
To illustrate this, in our experiments, we generate two additional invisible cloaks that printed with another two types of adversarial patches, where YOLO v2 detector will recognize them as ``Kite'' and some other objects (``Others''), respectively.
Besides, we directly print the original images (``Leaves'' and ``Apples'') of these two additional invisible cloaks to launch the attacks, and compare their attack performances with the two generated invisible cloaks, so as to demonstrate the necessity of the proposed adversarial method.
The orange teddy bear image (as shown in Figure \ref{fig:fig5}(a)) is used to generate these six ``teddy bear'' adversarial invisible cloaks that presented in Figure \ref{fig:fig8}(a)-\ref{fig:fig8}(f).
The green leaves image (Figure \ref{fig:fig8}(i)) and the red apples image (Figure \ref{fig:fig8}(j)) are used to generate the adversarial ``Kite-cloak'' (Figure \ref{fig:fig8}(g)) and the adversarial ``Others-cloak'' (Figure \ref{fig:fig8}(h)), respectively.
The printed cloaks are shown in Figure \ref{fig:fig8}.

To compare with these six ``teddy bear'' invisible cloaks, for those cloaks that printed with other types of patches, we take 100 photos for the testers in the same indoor and outdoor physical scenes of (3m, $15^\circ$).
The results show that, without the adversarial transformations, the three clothes printed with the original images (as shown in Figure \ref{fig:fig8}(i)-\ref{fig:fig8}(k)) are completely ineffective, and their stealth attack success rates are 0\%.
However, once the proposed adversarial transformations have been performed on these original images, their generated invisible cloaks (as shown in Figure \ref{fig:fig8}(a)-\ref{fig:fig8}(h)) will be effective for the person stealth attacks.
Specifically, the average attack performance of six ``teddy bear'' invisible cloaks is high up to 94\%. The Kite-cloak and Others-cloak generated from the green leaves and red apples achieve the stealth attack success rate of 28\% and 52\% in the physical world, respectively.
This indicates that the adversarial transformations are necessary, and the proposed stealth attack is generalizable.

\begin{figure}[htbp]
  \centering
  \includegraphics[width=1.9in]{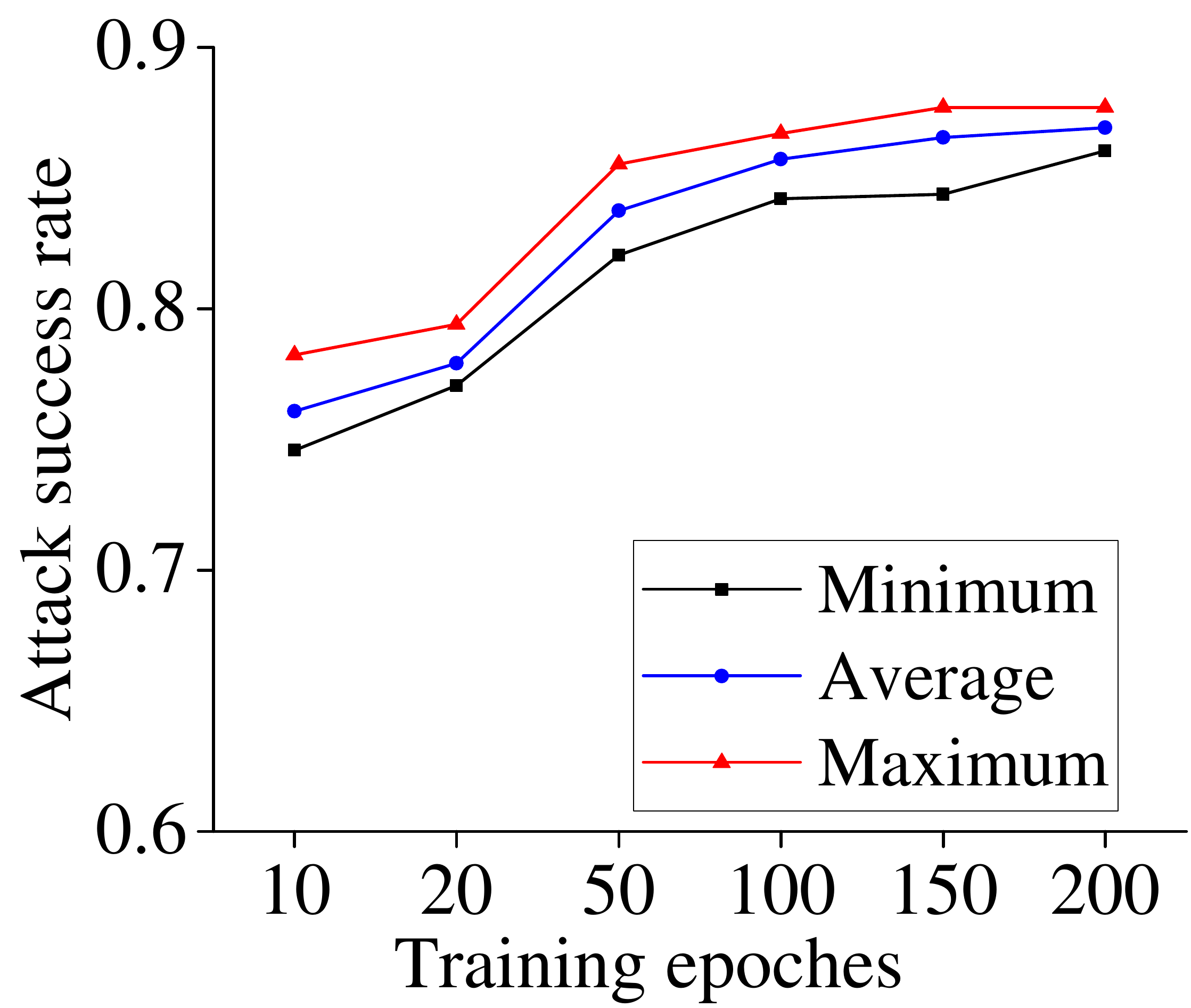}\\
  \caption{Attack performance of the combined-patch under different training epoches.}
  \label{fig:fig10}
\end{figure}

\textbf{Convergence of the proposed method.} To demonstrate the convergence of the proposed person stealth attack, we evaluate its performance in digital domain (Inria dataset) under different training epoches.
As discussed in Section \ref{sec:sec6.2}, we repeat the attacks for 10 times and report their maximum, minimum, and average attack success rate, respectively.
Figure \ref{fig:fig10} shows the attack performance of Combined-patch under different training epoches.
It is shown that, as the number of training epoches increases from 10 to 50, the maximum, minimum and average attack success rate of Combined-patch has been significantly improved.
The proposed person stealth attack method will converge after a certain number of epoches (around 150).
In other words, the maximum, minimum and average attack performance of proposed method all will be stable.
Specifically, after 150 training epoches, the attack success rate of the generated Combined-patch in digital domain (Inria dataset \cite{dalal2005histograms}) is 84.38\% (minimum), 87.70\% (maximum) and 86.56\% (average), respectively.

\vspace{-0.25cm}

\end{document}